\documentclass[10pt,twocolumn,letterpaper]{article}

\usepackage[final]{cvpr}      

\usepackage{graphicx}
\usepackage{amsmath}
\usepackage{amssymb}
\usepackage{booktabs}
\usepackage{multirow}
\usepackage{array}
\usepackage{bm}
\usepackage{mathrsfs}
\usepackage{algorithm}
\usepackage{algorithmic}
\usepackage{subfloat}

\usepackage[pagebackref,breaklinks,colorlinks]{hyperref}

\usepackage[capitalize]{cleveref}
\crefname{section}{Sec.}{Secs.}
\Crefname{section}{Section}{Sections}
\Crefname{table}{Table}{Tables}
\crefname{table}{Tab.}{Tabs.}

\def\cvprPaperID{8566} 
\def\confName{CVPR}
\def\confYear{2022}

\begin{document}

\title{Sparse Local Patch Transformer for Robust Face Alignment and Landmarks Inherent Relation Learning}

\author{Jiahao Xia$^{1}$, Weiwei Qu$^{2}$, Wenjian Huang$^2$, Jianguo Zhang\thanks{Corresponding Author} $^{ ,2}$, Xi Wang$^3$, Min Xu\footnotemark[1] $^{, 1}$\\
{\tt\small Jiahao.Xia@student.uts.edu.au, 11930667@mail.sustech.edu.cn, huangwj@sustech.edu.cn,} \\
{\tt\small zhangjg@sustech.edu.cn, Xi.Wang@calmcar.com, Min.Xu@uts.edu.au}\\
{\footnotesize $^1$University of Technology Sydney, $^2$Southern University of Technology and Science, $^3$CalmCar}\\
}
\maketitle

\begin{abstract}
   
   Heatmap regression methods have dominated face alignment area in recent years while they ignore the inherent relation between different landmarks. In this paper, we propose a Sparse Local Patch Transformer (SLPT) for learning the inherent relation. The SLPT generates the representation of each single landmark from a local patch and aggregates them by an adaptive inherent relation based on the attention mechanism. The subpixel coordinate of each landmark is predicted independently based on the aggregated feature. Moreover, a coarse-to-fine framework is further introduced to incorporate with the SLPT, which enables the initial landmarks to gradually converge to the target facial landmarks using fine-grained features from dynamically resized local patches. Extensive experiments carried out on three popular benchmarks, including WFLW, 300W and COFW, demonstrate that the proposed method works at the state-of-the-art level with much less computational complexity by learning the inherent relation between facial landmarks. The code is available at the project website\footnote{\url{https://github.com/Jiahao-UTS/SLPT-master}}.
   
\end{abstract}

\section{Introduction}
\label{sec:intro}

Face alignment is aimed at locating a group of pre-defined facial landmarks from images. Robust face alignment based on deep learning technology has attracted increasing attention in recent years and it is the fundamental algorithm in many face-related applications such as face reenactment \cite{FreeNet}, face swapping \cite{AHF} and driver fatigue detection \cite{Fatigue}. Despite recent progress, it still remains a challenging problem, especially for images with heavy occlusion, profile view and illumination variation.

\begin{figure}[t!]
	\centering
	\includegraphics[width=\linewidth]{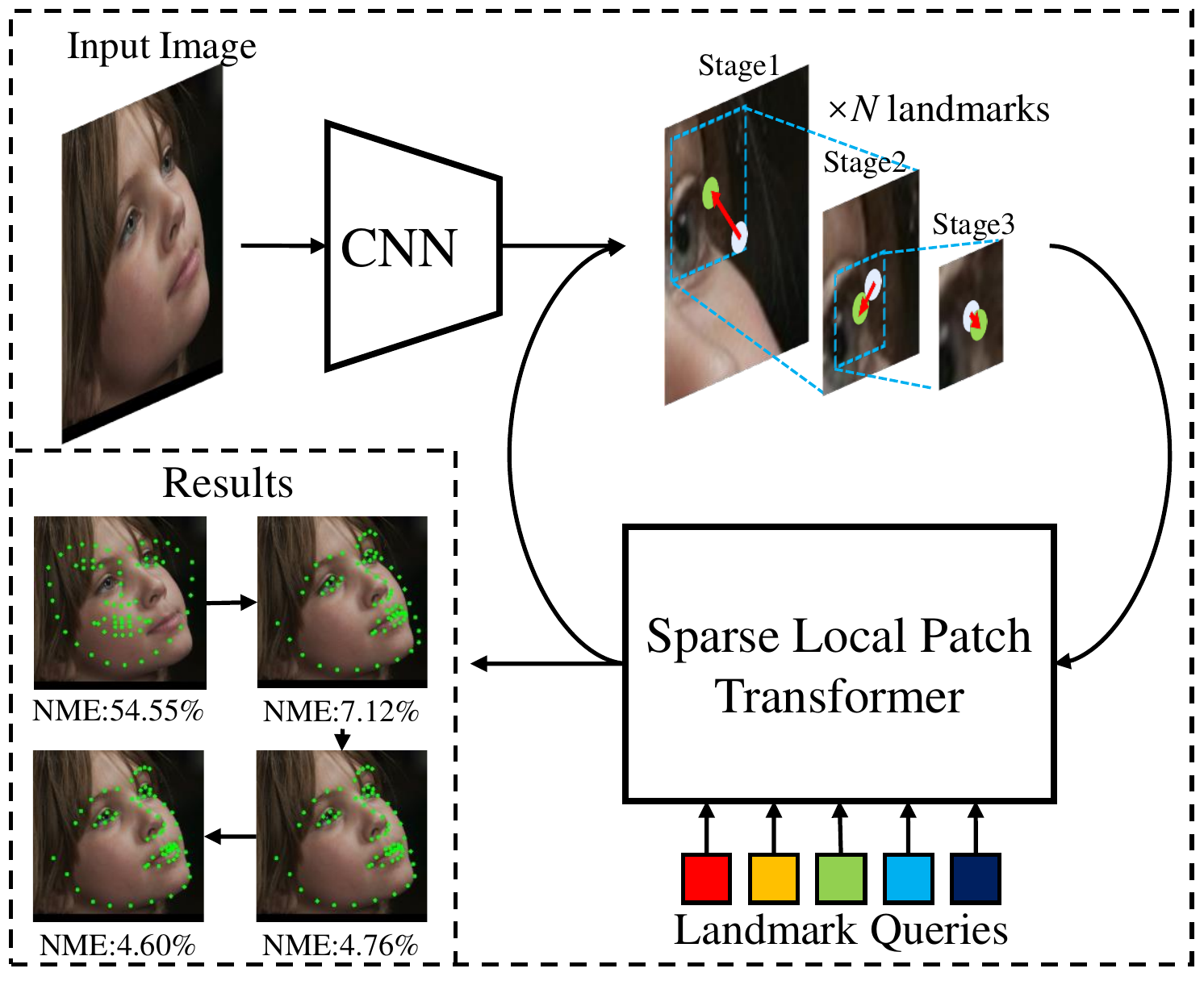}
	\caption{The proposed coarse-to-fine framework leverages the sparse local patches for robust face alignment. The sparse local patches are cropped according to the landmarks in the previous stage and fed into the same SLPT to predict the facial landmarks. Moreover, the patch size narrows down with the increasing of stages to enable the local features to evolve into a pyramidal form.}
	\label{fig1}
\end{figure}

The inherent relation between facial landmarks play an important role in face alignment since human face has a regular structure. Although heatmap regression methods achieve impressive performance \cite{MOHP, LUVLI, Awing, DeCaFA, HRnet} in recent years, they still ignore the inherent relation because convolutional neural network (CNN) kernels focus locally, thus failed to capture the relations of landmarks farther away in a global manner. In particular, they consider the pixel coordinate with highest intensity of the output heatmap as the optimal landmark, which inevitably introduces a quantization error, especially for common downsampled heatmap. Coordinate regression methods \cite{SCDF, SAN, Wing, TCDNN, LAB, DVLN, SRT} have an innate potential to learn the relation since it regresses the coordinates from global feature directly via fully-connected layers (FC). Nevertheless, a coherent relation should be learned together with local appearance while coordinate regression methods lose the local feature by projecting the global feature into FC layers.

To address the aforementioned problems, we propose a Sparse Local Patch Transformer (SLPT). Instead of predicting the coordinates from the full feature map like DETR \cite{DETR}, the SLPT firstly generates the representation for each landmark from a local patch. Then, a series of learnable queries, which are called \textit{landmark queries}, are used to aggregate the representations. Based on the cross-attention mechanism of transformer, the SPLT learns an adaptive adjacency matrix in each layer. Finally, the subpixel coordinate of each landmark in their corresponding patch is predicted independently by a MLP. Due to the use of sparse local patches, the number of the input token decreases significantly compared to other vision transformer\cite{DETR, VIT}.

To further improve the performance, a coarse-to-fine framework is introduced to incorporate with the SLPT, as shown in Fig.1. Similar to cascaded shape regression method \cite{CFSS, DAN, DAC-CSR}, the proposed framework optimizes a group of initial landmarks to the target landmarks by several stages. The local patches in each stage are cropped based on the initial landmarks or the landmarks predicted in the former stage, and the patch size for a specific stage is $1/2$ of its former stage. As a result, the local patches evolve in a pyramidal form and get closer to the target landmarks for the fine-grained local feature.

To verify the effectiveness of the SLPT and the proposed framework, we carry out experiments on three popular benchmarks, WFLW\cite{LAB}, 300W\cite{300W} and COFW\cite{COFW}. The results show the proposed method significantly outperforms other state-of-the-art methods in terms of diverse metrics with much lower computational complexity. Moreover, we also visualize the attention map of SLPT and the inner product matrix of landmark queries to demonstrate the SLPT can learn the inherent relation of facial landmarks. 

The main contributions of this work can be summarized as:

\begin{itemize}
	\item We introduce a novel transformer, Sparse Local Patch Transformer, to explore the inherent relation between facial landmarks based on the attention mechanism. The adaptive inherent relation learned by SLPT enables the model to achieve SOTA performance with much less computational complexity.
	\item We introduce a coarse-to-fine framework to incorporate with the SLPT, which enables the local patch to evolve in a pyramidal form and get closer to the target landmark for the fine-grained feature.
	\item Extensive experiments are conducted on three popular benchmarks, WFLW, 300W and COFW. The result illustrates the proposed method learns the inherent relation of facial landmarks by the attention mechanism and works at the SOTA level. 
\end{itemize}

\section{Related Work}

In the early stage of face alignment, the mainstream methods \cite{CLM, SDM, Face3000, CFSS, SCDF, DAC-CSR, COFW, DCFE} regress facial landmarks directly from the local feature with classical machine learning algorithms like random forest. With the development of CNN, the CNN-based face alignment methods have achieved impressive performance. They can be roughly divided into two categories: heatmap regression method and coordinate regression method. 

\subsection{Coordinate Regression Method}
 Coordinate regression methods \cite{TCDNN, MTCNN, DVLN, Wing} regress the coordinates of landmarks from feature map directly via FC layers. To further improve the robustness, diverse cascaded networks \cite{MDM, DAN} and recurrent networks \cite{RAR} are proposed to achieve face alignment with multi stages. Despite coordinate regression methods have an innate potential to learn the inherent relation, it commonly requires a huge number of samples for training. To address the problem, Qian et al. \cite{AVS} and Dong et al. \cite{SAN} expand the number of training samples by style transfer; Browatzki et al. \cite{3FabRec} and Dong et al. \cite{SRT} leverage the unlabeled dataset to train the model. In recent years, state-of-the-art works employ the structure information of face as the prior knowledge for better performance. Lin et al. \cite{SCDF} and Li et al. \cite{SDL} model the interaction between landmarks by a graph convolutional network (GCN). However, the adjacency matrix of GCN is fixed during inference and cannot adjust case by case. Learning an \textit{adaptive} inherent relation is crucial for robust face alignment. Unfortunately, there is no work yet on this topic, and we propose a method to fill this gap.
 
 
 
\begin{figure*}[t!]
	\centering
	\includegraphics[width=\linewidth]{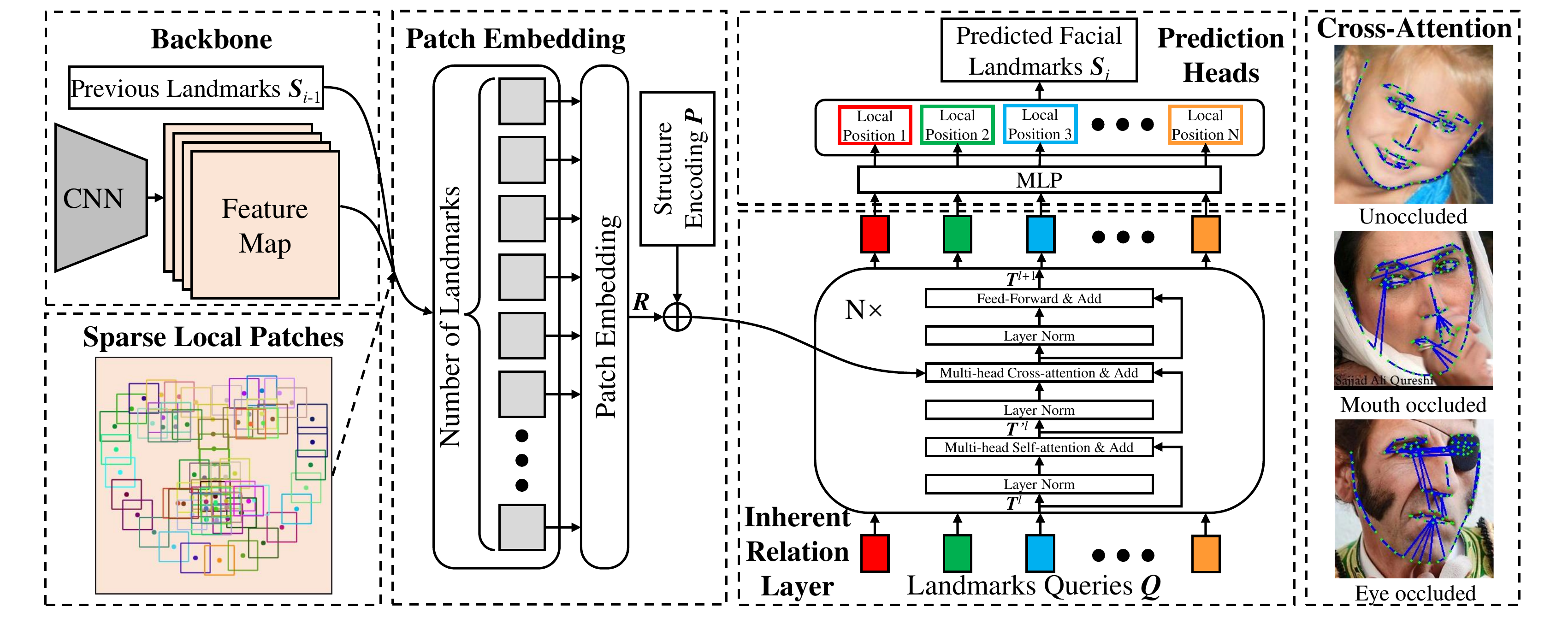}
	\caption{An overview of the SLPT. The SLPT crops local patches from the feature map according to the facial landmarks in the previous stage. Each patch is then embedded into a vector that can be viewed as the representation of the corresponding landmark. Subsequently, they are supplemented with the structure encoding to obtain the relative position in a regular face. A fixed number of landmark queries are then input into the decoder, attending the vectors to learn the inherent relation between landmarks. Finally, the outputs are fed into a shared MLP to estimate the position of each facial landmark independently. The rightmost images demonstrate the adaptive inherent relation of different samples. We connect each point to the point with highest cross-attention weight in the first inherent relation layer.}
	\label{fig2}
\end{figure*}
\subsection{Heatmap Regression Method}
Heatmap regression methods \cite{Hourglass, Dunet, HRnet, DeCaFA} output an intermediate heatmap for each landmark and consider the pixel with highest intensity as the optimal output. Therefore, it leads to quantization errors since the heatmap is commonly much smaller than the input image. To eliminate the error, Kumar et al. \cite{LUVLI} estimate the uncertainty of predicted landmark locations; Lan et al \cite{HIH} adopt an additional decimal heatmap for subpixel estimation; Huang et al. \cite{ADNet} further regress the coordinate from an anisotropic attention mask generated from heatmaps. Moreover, heatmap regression methods also ignore the relation between landmarks. To construct the relation between neighboring points, Wu et al. \cite{LAB} and Wang et al. \cite{Awing} take advantage of facial boundaries as the prior knowledge; Zou et al. \cite{HLSE} cluster landmarks with a graph model to provide structural constraints. However, they still cannot explicitly model an inherent relation between the landmarks with long distance.

The vision transformer \cite{VIT} proposed recently enables the model to attend the area with a long distance. Besides, the attention mechanism in transformer can generate an adaptive global attention for different tasks, such as object detection \cite{DETR, Deformable_DETR} and human pose estimation \cite{TokenPose}, and in principle, we envision that it can also learn an adaptive inherent relation for face alignment. In this paper, we demonstrate the capability of SLPT for learning the relation. 


\section{Method}
\subsection{Sparse Local Patch Transformer}
As shown in Fig.2, Sparse Local Patch Transformer (SLPT) consists of three parts, the patch embedding \& structure encoding, inherent relation layers and prediction heads.

\textbf{Patch embedding \& structure encoding}: ViT \cite{VIT} divides an image or a feature map $\bm{I} \in \mathbb{R}^{H_I \times W_I \times C}$ into a grid of $\frac{H_I}{P_h} \times \frac{W_I}{P_w}$ with each patch of size $P_h \times P_w$ and maps it into a $d$-dimension vector as the input. Different from ViT, for each landmark, the SLPT crops a local patch with the fixed size $\left( P_h, P_w \right)$ from the feature map as its supporting patch, whose center is located at the landmark. Then, the patches are resized to $K \times K$ by linear interpolation and mapped into a series of vectors by a CNN layer. Hence, each vector can be viewed as the representation of the corresponding landmark. Besides, to retain the relative position of landmarks in a regular face shape (structure information), we supplement the representations with a series of learnable parameters called \textit{structure encoding}. As shown in Fig.3, the SLPT learns to encode the distance between landmarks within the regular facial structure in the similarity of encodings. Each encoding has high similarity with the encoding of neighboring landmark (eg. left eye and right eye).


\textbf{Inherent relation layer}: Inspired by Transformer \cite{Transformer}, we propose inherent relation layers to model the relation between landmarks. Each layer consists of three blocks, multi-head self-attention (MSA) block, multi-head cross-attention (MCA) block, and multilayer perceptron (MLP) block, and an additional Layernorm (LN) is applied before every block. Based on the self-attention mechanism in MSA block, the information of queries interact adaptively for learning a $query-query$ inherent relation. Supposing the $l$-th MSA block obtains $H$ heads, the input $T^l$ and landmark queries $Q$ with $C_I$-dimension are divided into $H$ sequences equally ($T^l$ is a zero matrix in $1$st layer). The self-attention weight of the $h$-th head $\bm{A}_h$ is calculated by:
\begin{equation}
	\bm{A}_h=softmax\left ( \frac{\left(\bm{T}_h^{l} + \bm{Q}_h\right) \bm{W}^q_h \left( \left(\bm{T}_h^{l} + \bm{Q}_h\right)\bm{W}^k_h\right)^T}{\sqrt{C_h}}\right ),
\end{equation}
where $\bm{W}^q_h$ and $\bm{W}^k_h$ $\in \mathbb{R} ^ {C_h \times C_h}$ are the learnable parameters of two linear layers. $\bm{T}^{l}_{h} \in \mathbb{R} ^{N \times C_h}$ and $\bm{Q}_h \in \mathbb{R} ^{N \times C_h}$ are the input and landmark queries respectively of the $h$-th head with the dimension $C_h=C_I/H$. Then, MSA block can be formulated as:
\begin{equation}
	MSA\left(\bm{T}^{l} \right) = \left[ \bm{A}_1\bm{T}^l_1\bm{W}^v_1;...;\bm{A}_H\bm{T}^l_H\bm{W}^v_H \right]\bm{W}_P,
\end{equation}
where $\bm{W}_h^v \in \mathbb{R}^{C_h \times C_h}$ and $\bm{W}_P \in \mathbb{R}^{C_I \times C_I}$ are also the learnable parameters of linear layers. 

The MCA block aggregates the representations of facial landmarks based on the cross-attention mechanism for learning an adaptive $representation-query$ relation. As shown in the rightmost images of Fig.2, by taking advantage of the cross attention, each landmark can employ neighboring landmarks for coherent prediction and the occluded landmark can be predicted according to the representations of visible landmarks. Similar to MSA, MCA also has $H$ heads and the attention weight in the $h$-th head $\bm{A}_h^\prime$ can be calculated by:
\begin{equation}
	\bm{A}_h^\prime=softmax\left ( \frac{\left(\bm{T}_h^{\prime l} + \bm{Q}_h\right) \bm{W}^{\prime q}_h \left( \left(\bm{R}_h + \bm{P}_h\right)\bm{W}^{\prime  k}_h\right)^T}{\sqrt{C_h}}\right ).
\end{equation}
Where $\bm{W}^{\prime q}_h$ and $\bm{W}^{\prime k}_h \in \mathbb{R} ^ {C_h \times C_h}$ are learnable parameters of two linear layers in the $h$-th head. $\bm{T}_h^{\prime l} \in \mathbb{R} ^{N \times C_h}$ is the input $l$-th MCA block; $\bm{P}_h \in \mathbb{R} ^{N \times C_h}$ is the structure encodings; $\bm{R}_h \in \mathbb{R} ^{N \times C_h}$ is the landmark representations. MCA block can be formulated as:
\begin{equation}
	MCA\left(\bm{T}^{\prime l} \right) = \left[ \bm{A}^{\prime}_1\bm{T}^{\prime l}_1\bm{W}^{\prime v}_1;...;\bm{A}^{\prime}_H\bm{T}^{\prime l}_H\bm{W}^{\prime v}_H \right]\bm{W}^{\prime}_P,
\end{equation}
where $\bm{W}_h^{\prime v} \in \mathbb{R}^{C_h \times C_h}$ and $\bm{W}^{\prime}_P \in \mathbb{R}^{C_I \times C_I}$ are also the learnable parameters of linear layers in MCA block.

Supposing predicting $N$ pre-defined landmarks, the computational complexity of the MCA that employ sparse local patches $\Omega(S)$ and full feature map $\Omega(F)$ is:
\begin{figure}[t!]
	\centering
	\includegraphics[width=\linewidth]{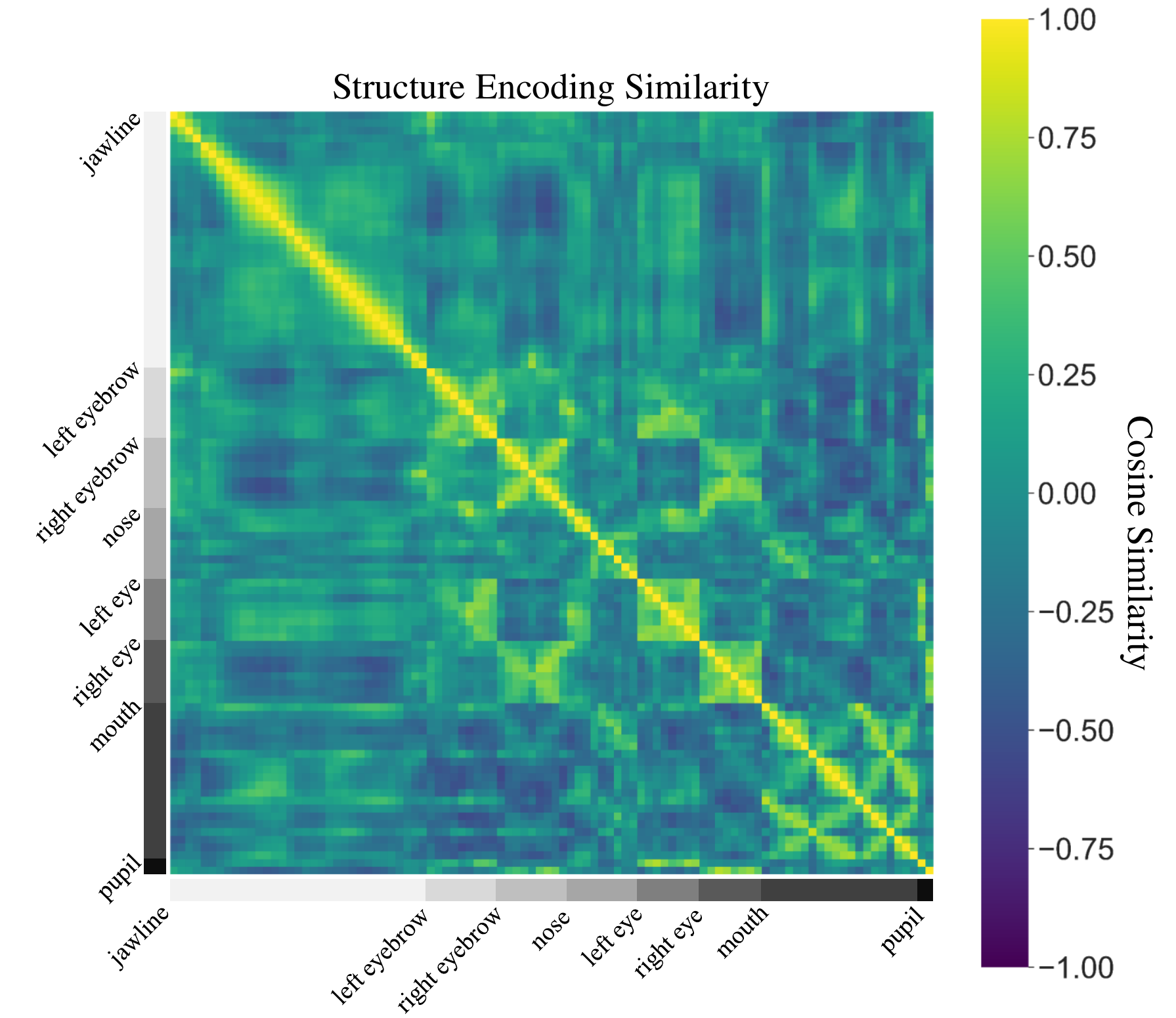}
	\caption{Cosine similarity for structure encodings of SLPT learned from a dataset with 98 landmark annotations. High cosine similarities are observed for the corresponding points which are close in the regular face structure.}
	\label{fig3}
\end{figure}
\begin{equation}
	\Omega(S)=4HNC_h^2 + 2HN^2C_h,
\end{equation}
\begin{equation}
	\Omega(F)=\left(2N+2\frac{W_IH_I}{P_wP_h} \right)HC_h^2 + 2NH\frac{W_IH_I}{P_wP_h}C_h.
\end{equation}
Compared to using the full feature map, the number of representations decreases from $\frac{H_I}{P_h} \times \frac{W_I}{P_w}$ to $N$ (with the same input size, $\frac{H_I}{P_h} \times \frac{W_I}{P_w}$ is $16 \times 16$ in the related framework \cite{DETR}), which decreases the computational complexity significantly. For a 29 landmark dataset \cite{COFW},  $\Omega(S)$ is only $1/5$ of $\Omega(F)$ ($H = 8$ and $C_h = 32$ in the experiment).

\textbf{Prediction head}: the prediction head consists of a layernorm to normalize the input and a MLP layer to predict the result. The output of the inherent relation layer is the local position of the landmark with respect to its supporting patch. Based on the local position on the $i$-th patch $\left(t_x^i, t_y^i\right)$, the global coordinate of the $i$-th landmark $\left( x^i, y^i\right)$ can be calculated by:
\begin{equation}
\begin{aligned}
	x^i &= x^i_{lt} + w^i t^i_x,\\
	y^i &= y^i_{lt} + h^i t^i_y,
\end{aligned}
\end{equation}
where $(w^i, h^i)$ is the size of the supporting patch.

\begin{algorithm}[t!]
	\caption{Training pipeline of the coarse-to-fine framework}
	\label{alg1}
	\begin{algorithmic}[1]
		\REQUIRE Training image $\bm{I}$, initial landmarks $\bm{S}_0$, backbone network $B$, SLPT $T$, loss function $L$, ground truth $\bm{S}_{gt}$, Stage number $N_{stage}$
		\WHILE{the training epoch is less than a specific number}
		\STATE Forward $B$ for feature map by $\bm{F}=B\left( I\right)$;
		\STATE Initialize the local patch size $\left(P_w, P_h\right) \leftarrow \left(\frac{W}{4}, \frac{H}{4}\right)$
		\FOR{ $i$ $\leftarrow$ 1 \TO $N_{stage}$ }
		\STATE Crop local pactes $\bm{P}$ from $\bm{F}$ according to former landmarks $\bm{S}_{i-1}$;
		\STATE Resize patches from $\left(P_w, P_h\right)$ to $K \times K$;
		\STATE Forward $T$ for landmarks by $\bm{S}_{i}=T\left( \bm{P} \right)$;
		\STATE Reduce the patch size $\left(P_w, P_h\right)$ by half;
		\ENDFOR
		\STATE Minimize $L\left(\bm{S_{gt}}, \bm{S_{1}}, \bm{S_{2}}, \cdots , \bm{S}_{N_{stage}} \right)$
		\ENDWHILE
	\end{algorithmic} 
\end{algorithm}

\subsection{Coarse-to-fine locating}
To further improve the performance and robustness of SLPT, we introduce a coarse-to-fine framework trained in an end-to-end method to incorporate with the SLPT. The pseudo-code in \textbf{Algorithm 1} shows the training pipeline of the framework. It enables a group of initial facial landmarks $\bm{S}_0$ calculated from the mean face in the training set to converge to the target facial landmarks gradually with several stages. Each stage takes the previous landmarks as center to crop a series of patches. Then, the patches are resized into a fixed size $K \times K$ and fed into the SLPT to predict the local point on the supporting patches. Large patch size in the initial stage enables the SLPT to obtain a large receptive filed that prevents the patch from deviating from the target landmark. Then, the patch size in the following stages is $1/2$ of its former stage, which enables the local patches to extract fine-grained features and evolve into a pyramidal form. By taking advantage of the pyramidal form, we can observe a significant improvement for SLPT. (see Section 4.5).

\begin{table}[t!]
	\centering
	\begin{tabular}{m{2.3cm}<{\centering}|m{1.3cm}<{\centering}m{1.3cm}<{\centering}m{1.3cm}<{\centering}}
		\hline
		Method & NME(\%)$\downarrow$ & FR$_{0.1}$(\%)$\downarrow$ & AUC$_{0.1}$$\uparrow$\\ \hline
		LAB \cite{LAB} & 5.27 & 7.56 & 0.532 \\
		SAN \cite{SAN} & 5.22 & 6.32 & 0.535 \\
		Coord$^\star$ \cite{HRnet} & 4.76 & 5.04 & 0.549 \\
		DETR$^\dag$ \cite{DETR} & 4.71 & 5.00 & 0.552 \\
		Heatmap$^\star$ \cite{HRnet} & 4.60 & 4.64 & 0.524 \\
		AVS+SAN \cite{AVS} & 4.39 & 4.08 & 0.591 \\
		LUVLi \cite{LUVLI} & 4.37 & 3.12 & 0.557 \\
		AWing \cite{Awing} & 4.36 & 2.84 & 0.572 \\
		SDFL$^\star$ \cite{SCDF} & 4.35 & {\color{red}$\bm{2.72}$} & 0.576 \\
		SDL$^\star$ \cite{SDL} & 4.21 & 3.04 & 0.589 \\
		HIH \cite{HIH} & 4.18 & 2.84 & {\color{blue}$\bm{0.597}$} \\ 
		ADNet \cite{ADNet} & {\color{red}$\bm{4.14}$} & {\color{red}$\bm{2.72}$} & {\color{red}$\bm{0.602}$} \\ \hline
		SLPT$^\ddag$ & {\color{blue} $\bm{4.20}$} & 3.04 & 0.588 \\
		SLPT$^\dag$ & {\color{red}$\bm{4.14}$} & {\color{blue} $\bm{2.76}$} & 0.595 \\ \hline
	\end{tabular}
	\caption{Performance comparison of the SLPT and the state-of-the-art methods on WFLW. The normalization factor is inter-ocular and the threshold for FR is set to 0.1. Key: [{\color{red} \textbf{Best}}, {\color{blue} \textbf{Second Best}}, $^\star$=HRNetW18C, $^\dag$=HRNetW18C-lite, $^\ddag$=ResNet34]}
	\label{Tabal1}
\end{table}

\subsection{Loss Function}
We employ the normalized L2 loss to provide the supervision for stages of the coarse-to-fine framework. Moreover, similar to other works \cite{Hourglass, Dunet}, providing additional supervision for the intermediate output during the training is also helpful. Therefore, we feed the intermediate output of each inherent relation layer into a shared prediction head. The loss function is written as:
\begin{equation}
	L = \frac{1}{SDN} \sum_{i=1}^{S} \sum_{j=1}^{D} \sum_{k=1}^{N} \frac{\left\| \left( x_{gt}^k, y_{gt}^k\right) - \left( x^{ijk}, y^{ijk}\right) \right\|_2}{d},
\end{equation}
where $S$ and $D$ indicate the number of coarse-to-fine stage and inherent relation layer respectively. $\left(x_{gt}^k, y_{gt}^k\right)$ is the labeled coordinate of the $k$-th point. $\left(x^{ijk}, y^{ijk}\right)$ is the coordinate of $k$-th point predicted by $j$-th inherent relation layer in $i$-th stage. $d$ is the distance between outer eye corners that acts as a normalization factor.

\section{Experiment}
\subsection{Datasets}
Experiments are conducted on three popular benchmarks, including WFLW \cite{LAB}, 300W\cite{300W} and COFW\cite{COFW}.

\textbf{WFLW} dataset is a very challenging dataset that consists of 10,000 images, 7,500 for training and 2,500 for testing. It provides 98 manually annotated landmarks and rich attribute labels, such as profile face, heavy occlusion, make-up and illumination.

\textbf{300W} is the most commonly used dataset that includes 3,148 images for training and 689 images for testing. The training set consists of the fullset of AFW \cite{AFW}, the training subset of HELEN \cite{HELEN} and LFPW \cite{LFPW}. The test set is further divided into a challenging subset that includes 135 images (IBUG fullset \cite{300W}) and a common subset that consists of 554 images (test subset of HELEN and LFPW). Each image in 300W is annotated with 68 facial landmarks.

\textbf{COFW} mainly consists of the samples with heavy occlusion and profile face. The training set includes 1,345 images and each image is provided with 29 annotated landmarks. The test set has two variants. One variant presents 29 landmarks annotation per face image (COFW),  The other is provided with 68 annotated landmarks per face image (COFW68 \cite{COFW68}). Both contains 507 images. We employ the COFW68 set for \textit{cross}-dataset validation. 

\begin{table}[t!]
	\begin{tabular}{m{2.2cm}<{\centering}|m{1.2cm}<{\centering}m{1.6cm}<{\centering}m{1.2cm}<{\centering}}
		\hline
		\multirow{2}{*}{Method} &  \multicolumn{3}{c}{Inter-Ocular NME (\%) $\downarrow$} \\
		&  Common & Challenging & Fullset \\ \hline
		SAN \cite{SAN} &  3.34 & 6.60 & 3.98 \\
		Coord$^\star$ \cite{HRnet} &  3.05 & 5.39 & 3.51 \\ 
		LAB \cite{LAB} &  2.98 & 5.19 & 3.49 \\
		DeCaFA \cite{DeCaFA} &  2.93 & 5.26 & 3.39 \\
		HIH \cite{HIH} & 2.93 & 5.00 & 3.33  \\
		Heatmap$^\star$ \cite{HRnet} & 2.87 & 5.15 & 3.32 \\
		SDFL$^\star$ \cite{SCDF} & 2.88 & 4.93 & 3.28 \\
		HG-HSLE \cite{HLSE} & 2.85 & 5.03 & 3.28 \\
		LUVLi \cite{LUVLI} & 2.76 & 5.16 & 3.23 \\
		AWing \cite{Awing} & 2.72 & {\color{red} \textbf{4.53}} & 3.07 \\ 
		SDL$^\star$ \cite{SDL} & {\color{blue} \textbf{2.62}} & 4.77 & {\color{blue} \textbf{3.04}} \\ 
		ADNet \cite{ADNet} & {\color{red} \textbf{2.53}} & {\color{blue} \textbf{4.58}} & {\color{red} \textbf{2.93}} \\ \hline
		SLPT$^\ddag$ & 2.78 & 4.93& 3.20 \\
		SLPT$^\dag$ & 2.75 & 4.90 & 3.17 \\ \hline
	\end{tabular}
	\caption{Performance comparison for SLPT and the state-of-the-art methods on 300W common subset, challenging subset and fullset. Key: [{\color{red} \textbf{Best}}, {\color{blue} \textbf{Second Best}}, $^\star$=HRNetW18C, $^\dag$=HRNetW18C-lite, $^\ddag$=ResNet34]}
	\label{Tabal2}
\end{table}
\begin{figure}[t!]
	\centering
	\includegraphics[width=\linewidth]{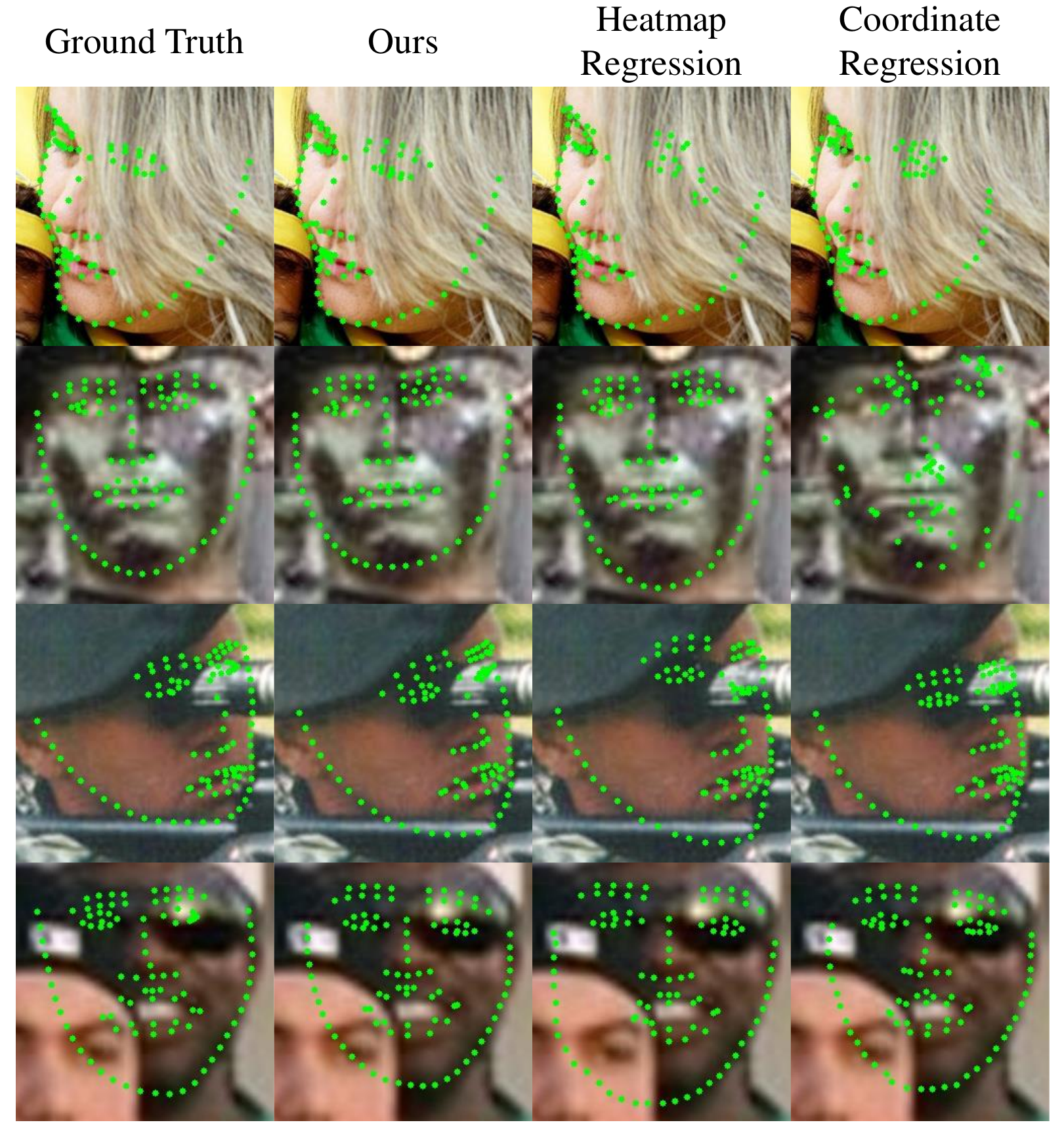}
	\caption{Visualization of the ground truth and face alignment result of SLPT, heatmap regression (HRNetW18C) and coordinate regression (HRNetW18C) method on the faces with blur, heavy occlusion and profile face.}
	\label{fig4}
\end{figure}

\subsection{Evaluation Metrics}
Referring to other related work \cite{LUVLI, Awing, SCDF}, we evaluate the proposed methods with standard metrics, Normalized Mean Error (NME), Failure Rate (FR) and Area Under Curve (AUC). \textbf{NME} is defined as:
\begin{equation}
	NME\left(\bm{S}, \bm{S}_{gt}\right) = \frac{1}{N}\sum_{i=1}^{N}\frac{\left\|\bm{p}^i-\bm{p}_{gt}^i\right\|_2}{d} \times 100\%,
\end{equation}
where $\bm{S}$ and $\bm{S}_{gt}$ denote the predicted and annotated coordinates of landmarks respectively. $\bm{p}^i$ and $\bm{p}^i_{gt}$ indicate the coordinate of $i$-th landmark in $\bm{S}$ and $\bm{S}_{gt}$. $N$ is the number of landmarks, $d$ is the reference distance to normalize the error. $d$ could be the distance between outer eye corners (inter-ocular) or the distance between pupil centers (inter-pupils). \textbf{FR} indicates the percentage of images in the test set whose NME is higher than a certain threshold. \textbf{AUC} is calculated based on Cumulative Error Distribution (CED) curve. It indicates the fraction of test images whose NME(\%) is less or equal to the value on the horizontal axis. AUC is the area under CED curve, from zero to the threshold for FR.

\begin{table}[t!]
	\centering
	\begin{tabular}{m{2.2cm}<{\centering}|m{1.05cm}<{\centering}m{0.95cm}<{\centering}|m{1.05cm}<{\centering}m{0.95cm}<{\centering}}
		\hline
		\multirow{2}{*}{Method} &  \multicolumn{2}{c|}{Inter-Ocular} & \multicolumn{2}{c}{Inter-Pupil} \\
		& NME(\%)$\downarrow$ & FR(\%)$\downarrow$ & NME(\%)$\downarrow$ & FR(\%)$\downarrow$ \\  \hline
		DAC-CSR \cite{DAC-CSR} & 6.03 & 4.73 & - & -\\
		LAB \cite{LAB} & 3.92 & 0.39 & - & -\\
		Coord$^\star$ \cite{HRnet} & 3.73 & 0.39 &- & -\\
		SDFL$^\star$ \cite{SCDF} & 3.63 & {\color{red} $\bm{0.00}$} & - & - \\
		Heatmap$^\star$ \cite{HRnet} & 3.45 & {\color{blue} $\bm{0.20}$} & - & -\\
		Human \cite{COFW} & -& - & 5.60 & - \\
		TCDCN \cite{TCDNN} & - & - & 8.05 & - \\
		Wing \cite{Wing} &-&- & 5.44 & 3.75 \\
		DCFE \cite{DCFE} &-&-& 5.27 & 7.29 \\
		AWing \cite{Awing} &-&-& 4.94 & {\color{blue} $\bm{0.99}$} \\
		ADNet \cite{ADNet} &-&-& {\color{red} $\bm{4.68}$} & {\color{red} $\bm{0.59}$} \\ \hline
		SLPT$^\ddag$  & {\color{blue} $\bm{3.36}$} & 0.59 & 4.85 &  1.18\\
		SLPT$^\dag$ & {\color{red} $\bm{3.32}$} & {\color{red} $\bm{0.00}$} & {\color{blue} $\bm{4.79}$} & 1.18 \\ \hline		
	\end{tabular}
	\caption{NME and FR$_{0.1}$ comparisons under Inter-Ocular normalization and Inter-Pupil normalization on $within$-dataset validation. The threshold for failure rate (FR) is set to 0.1. Key: [{\color{red} \textbf{Best}}, {\color{blue} \textbf{Second Best}}, $^\star$=HRNetW18C, $^\dag$=HRNetW18C-lite, $\ddag$=ResNet34]}
	\label{Tabal3}
\end{table}

\subsection{Implementation Details}
Each input image is cropped and resized to $256 \times 256$. We train the proposed framework with Adam \cite{Adam}, setting the initial learning rate to $1\times10^{-3}$. Without specifications, the size of the resized patch is set to $7 \times 7$ and the framework has 6 inherent relation layers and 3 coarse-to-fine stages. Besides, we augment the training set with random  horizontal flipping ($50\%$), gray ($20\%$), occlusion ($33\%$), scaling ($\pm5\%$), rotation ($\pm30^{\circ}$), translation ($\pm 10 px$). We implement our method with two different backbone: a light HRNetW18C \cite{HRnet} (the modularized block number in each stage is set to 1) and Resnet34\cite{Resnet}. For the HRNetW18C-lite, the resolution of feature map is $64 \times 64$, and for the Resnet34, we extract representations from the output feature maps of stages C2 through C5. (see Appendix A.1).

\begin{table}[t!]
	\centering
	\begin{tabular}{m{2.4cm}<{\centering}|m{1.8cm}<{\centering}m{1.6cm}<{\centering}}
		\hline
		Method & Inter-Pupil NME(\%)$\downarrow$ & FR$_{0.1}$(\%)$\downarrow$ \\ \hline
		TCDCN \cite{TCDNN} & 7.66 & 16.17 \\
		CFSS \cite{CFSS} & 6.28 & 9.07 \\
		ODN \cite{ODN} & 5.30 & - \\
		AVS+SAN \cite{AVS} & 4.43 & 2.82 \\
		LAB \cite{LAB} & 4.62 & 2.17 \\
		SDL$^\star$ \cite{SDL} & 4.22 & {\color{blue} $\bm{0.39}$} \\
		SDFL$^\star$ \cite{SCDF} & 4.18 & {\color{red} $\bm{0.00}$} \\ \hline
		SLPT$^\ddag$ & {\color{blue} $\bm{4.11}$} & 0.59  \\
		SLPT$^\dag$ & {\color{red} $\bm{4.10}$} & 0.59 \\ \hline
	\end{tabular}
	\caption{Inter-ocular NME and FR$_{0.1}$ comparisons on 300W-COFW68 \textit{cross}-dataset evaluation. Key: [{\color{red} \textbf{Best}}, {\color{blue} \textbf{Second Best}}, $^\star$=HRNetW18C, $^\dag$=HRNetW18C-lite, $^\ddag$=ResNet34]}
	\label{Tabal4}
\end{table}

\begin{table*}[t!]
	\centering
	\begin{tabular}{c|m{0.7cm}<{\centering}m{0.7cm}<{\centering}m{0.7cm}<{\centering}m{0.7cm}<{\centering}m{0.7cm}<{\centering}m{0.7cm}<{\centering}m{0.7cm}<{\centering}m{0.7cm}<{\centering}m{0.7cm}<{\centering}m{0.7cm}<{\centering}m{0.7cm}<{\centering}m{0.7cm}<{\centering}}
		\hline
		\multicolumn{1}{c|}{\multirow{3}{*}{Model}} & \multicolumn{12}{c}{Intermediate Stage}                                                                                                          \\ \cline{2-13} 
		\multicolumn{1}{c|}{} & \multicolumn{3}{c|}{1st stage} & \multicolumn{3}{c|}{2rd stage} & \multicolumn{3}{c|}{3rd stage} & \multicolumn{3}{c}{4th stage} \\ \cline{2-13} 
		\multicolumn{1}{c|}{} & NME & FR & \multicolumn{1}{c|}{AUC} & NME & FR & \multicolumn{1}{c|}{AUC} & NME & FR & \multicolumn{1}{c|}{AUC} & NME & FR & AUC \\ \hline
		Model$^\dag$ with 1 stage  &  4.79\%  &  5.08\%  & \multicolumn{1}{c|}{0.583}  & - &  - & \multicolumn{1}{c|}{-} & - & - & \multicolumn{1}{c|}{-}    & - &  -   &  - \\ \hline
		Model$^\dag$ with 2 stages & 4.52\% & 4.24\% & \multicolumn{1}{c|}{0.563} & 4.27\% & 3.40\% & \multicolumn{1}{c|}{0.585} & - & - & \multicolumn{1}{c|}{-}    & -  &  - & -  \\ \hline
		Model$^\dag$ with 3 stages  & 4.38\% & 3.60\%& \multicolumn{1}{c|}{0.574} & 4.16\%  & 2.80\%  & \multicolumn{1}{c|}{0.594} & {\color{red} $\bm{4.14\%}$} & {\color{red} $\bm{2.76\%}$} & \multicolumn{1}{c|}{{\color{red} $\bm{0.595}$}}    &  -  & -  &  - \\ \hline
		Model$^\dag$ with 4 stages & 4.47\% & 4.00\%  & \multicolumn{1}{c|}{0.567} & 4.26\% & 3.40\% & \multicolumn{1}{c|}{0.586} &4.24\% & 3.36\%  & \multicolumn{1}{c|}{0.588}    &  4.24\%  &  3.32\% & 0.587 \\ \hline
	\end{tabular}
	\caption{Performance comparison of the SLPT with different number of coarse-to-fine stages on WFLW. The normalization factor for NME is inter-ocular and the threshold for FR and AUC is set to 0.1. Key: [{\color{red} \textbf{Best}}, $^\dag$=HRNetW18C-lite]}
	\label{Tabel5}
\end{table*}

\subsection{Comparison with State-of-the-Art Method}
\textbf{WFLW}: as tabulated in Table 1 (more detailed results on the subset of WFLW are in Appendix A.2), SLPT demonstrates impressive performance. With the increasing of inherent layers, the performance of SLPT can be further improved and outperforms the ADNet (see Appendix A.5). Referring to DETR, we also implement a Transformer based method that employs the full feature map for face alignment. The number of the input tokens is $16 \times 16$. With the same backbone (HRNetW18C-lite), we observe an improvement of 12.10\% in NME, and the number of training epoch is $8 \times $ less than the DETR (see Appendix A.3). Moreover, the SLPT also outperforms the coordinate regreesion and heatmap regression methods significantly. Some qualitative results are shown in Fig. 4. It is evident that our method could localize the landmarks accurately, in particular for face images with blur (2$nd$ row in Fig.4), profile view (1$st$ row in Fig.4) and heavy occlusion (3$rd$ and 4$th$ row in Fig.4).

\textbf{300W}: the comparison result is shown in Table 2. Compared to the coordinate and heatmap regression methods (HRNetW18C \cite{HRnet}), SLPT still achieves an impressive improvement of 9.69\% and 4.52\% respectively in NME on the fullset. However, the improvement on 300W is not as significant as WFLW since learning an adaptive inherent relation requires a large number of annotated samples. With limited training samples, the methods with prior knowledge, such as facial boundaries (Awing and ADNet) and affined mean shape (SDL), always achieve better performance.

\begin{table}[t!]
	\centering
	\begin{tabular}{|m{1.4cm}<{\centering}|m{0.85cm}<{\centering}|m{0.85cm}<{\centering}|m{0.85cm}<{\centering}|m{0.85cm}<{\centering}|m{0.85cm}<{\centering}|}
		\hline
		Method & MSA & MCA & NME & FR & AUC \\ \hline
		Model$^\dag$ 1 &  w/o  & w/o  & 4.48\% & 4.32\% & 0.566 \\ \hline
		Model$^\dag$ 2 & w/ & w/o & 4.20\%  & 3.08\% & 0.590 \\ \hline
		Model$^\dag$ 3 & w/o & w/ & 4.17\% & 2.84\% & 0.593 \\ \hline
		Model$^\dag$ 4 & w/ & w/ & {\color{red} $\bm{4.14}$}\% & {\color{red} $\bm{2.76}$}\% & {\color{red} $\bm{0.595}$} \\ \hline
	\end{tabular}
	\caption{NME($\downarrow$), FR$_{0.1}$($\downarrow$) and AUC$_{0.1}$($\uparrow$) with/without Encoder and Decoder. Key: [{\color{red} \textbf{Best}}, $^\dag$=HRNetW18C-lite]}
	\label{Tabal6}
\end{table}

\begin{table}[t!]
	\centering
	\begin{tabular}{|m{3.4cm}<{\centering}|m{1.0cm}<{\centering}|m{1.0cm}<{\centering}|m{0.85cm}<{\centering}|}
		\hline
		Method & NME & FR & AUC \\ \hline
		w/o structure encoding$^\dag$ &  4.16\% & 2.84\% & 0.593 \\ \hline
		w structure encoding$^\dag$ &  {\color{red} $\bm{4.14}$}\% & {\color{red} $\bm{2.76}$}\% & {\color{red} $\bm{0.595}$} \\ \hline
	\end{tabular}
	\caption{NME($\downarrow$), FR$_{0.1}$($\downarrow$) and AUC$_{0.1}$($\uparrow$) with/without structure encoding. Key: [{\color{red} \textbf{Best}}, , $^\dag$=HRNetW18C-lite]}
	\label{Tabal7}
\end{table}

\textbf{COFW}: We conduct two experiments on COFW for comparsion, the $\textit{within}$-dataset validation and $\textit{cross}$-dataset validation. For the $\textit{within}$-dataset validation, the model is trained with 1,345 images and validated with 507 images on COFW. The inter-ocular and inter-pupil NME of SLPT and the state-of-the-art methods are reported in Table 3 respectively. In this experiment, the number of training sample is quite small, which leads to the significant degradation of the coordinate regression methods, such as SDFL, LAB. Nevertheless, SLPT still maintains excellent performance and yields the second best performance. It improves the metric by 3.77\% and 11.00\% in NME over the heatmap regression and coordinate regression methods respectively.

For the $\textit{cross}$-dataset validation, the training set includes the complete 300W dataset (3,837 images) and the test set is COFW68 (507 images with 68 landmark annotation). Most samples of COFW68 are under heavy occlusion. The inter-ocular NME and FR of SLPT and the state-of-the-art methods are reported in Table 4. Compared to the methods based on GCN (SDL and SDFL), the SLPT (HRNet) achieves impressive result, as low as 4.10\% in NME. The result illustrates that the adaptive inherent relation of SLPT works better than the fixed adjacency matrix of GCN for robust face alignment, especially for the condition of heavy occlusion.

\subsection{Ablation Study}

\begin{figure*}[t!]
	\label{verifycluster}
	\centering
	\subfloat[MCA-layer 1]{\includegraphics[width=2.7cm]{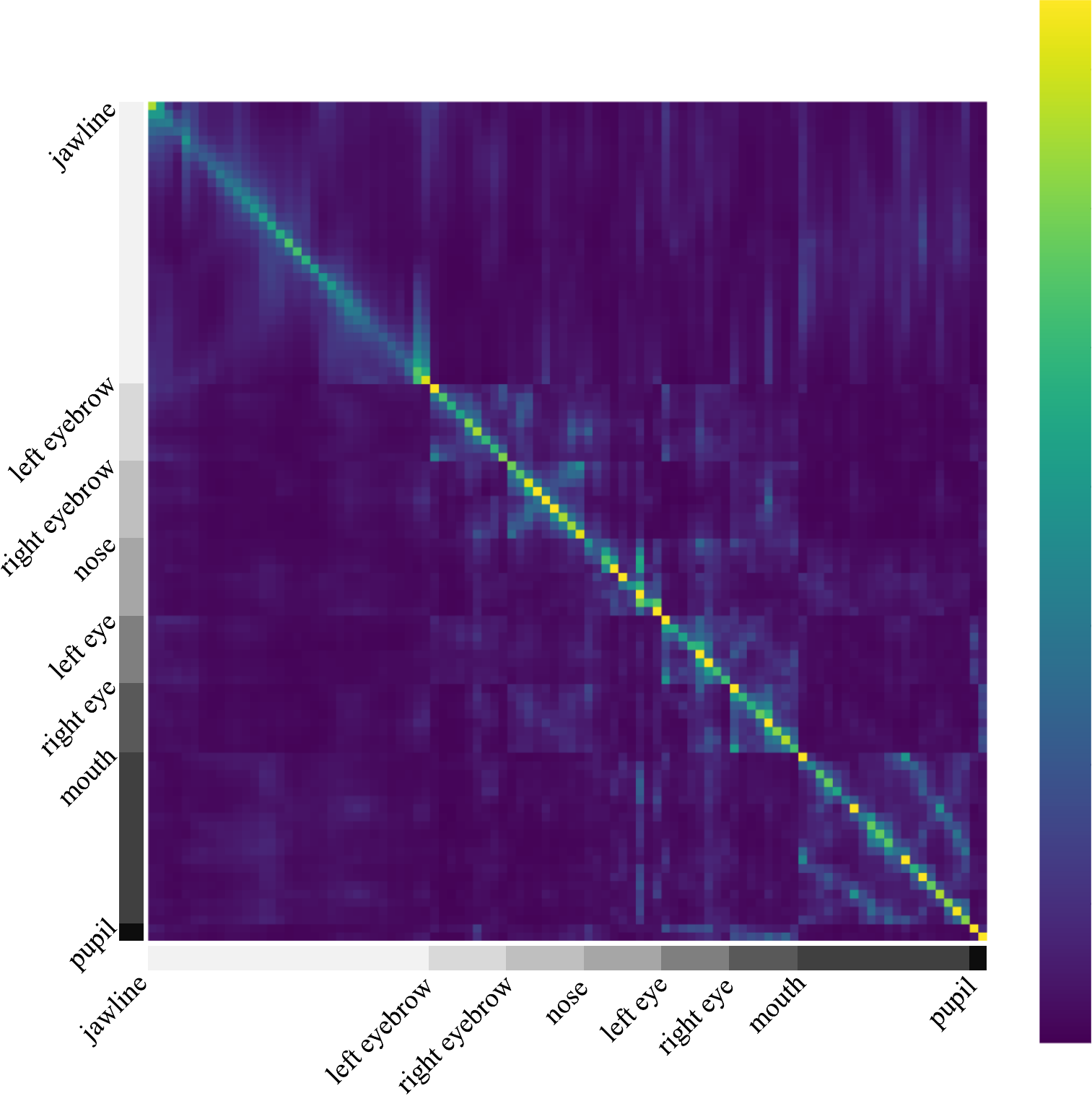}}\hspace{0.1cm}
	\subfloat[MCA-layer 2]{\includegraphics[width=2.7cm]{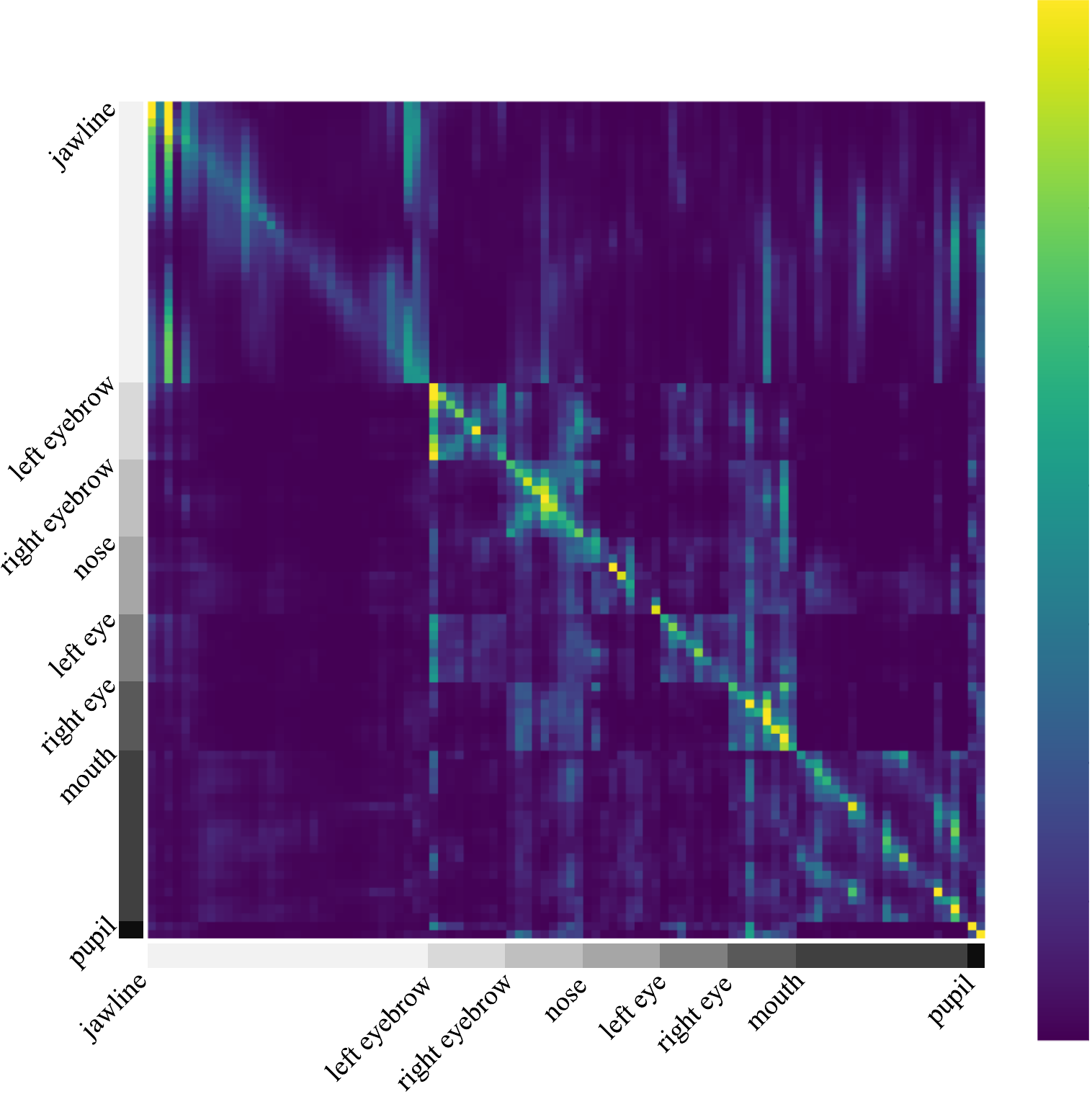}}\hspace{0.1cm}
	\subfloat[MCA-layer 3]{\includegraphics[width=2.7cm]{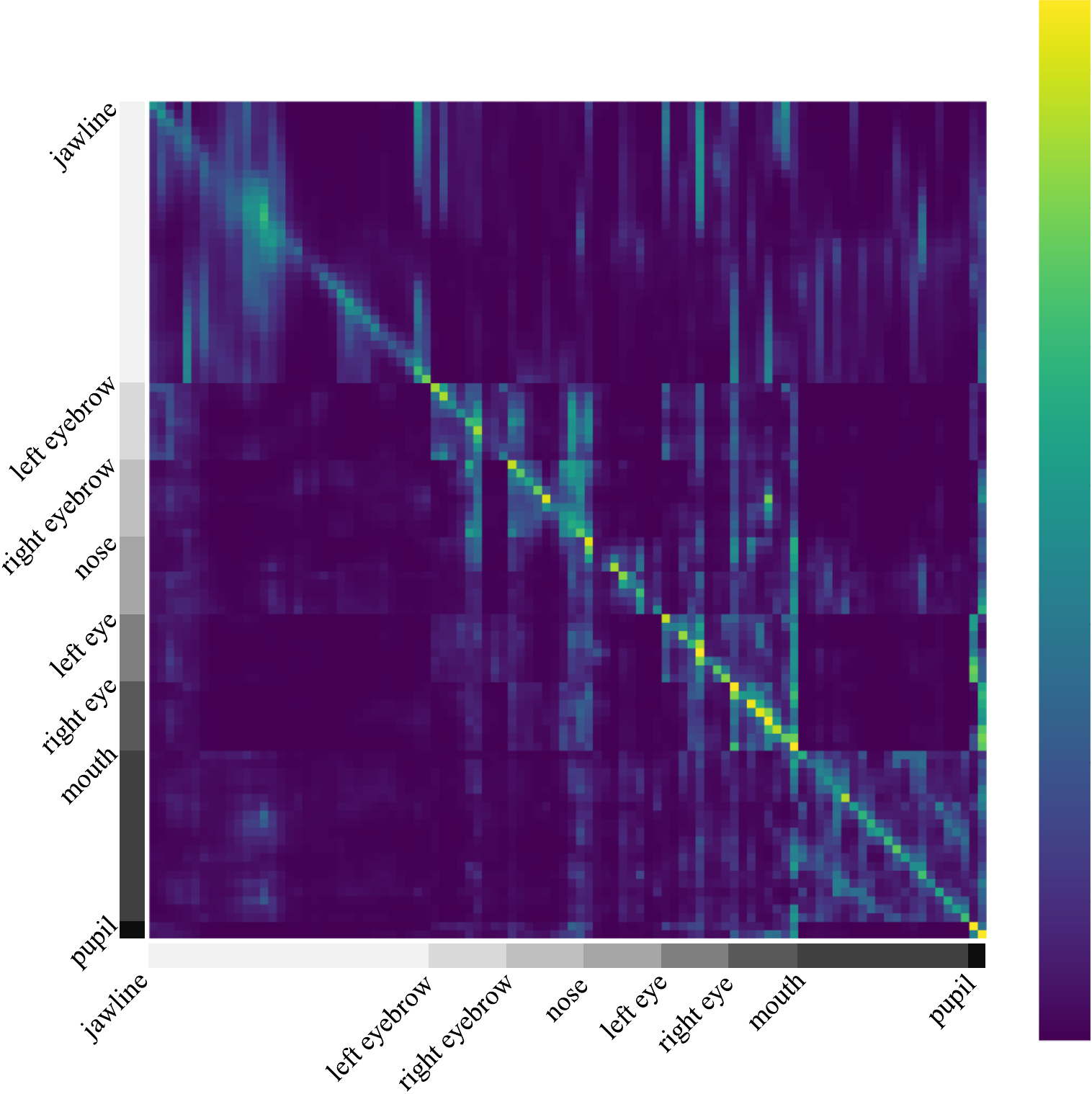}}\hspace{0.1cm}
	\subfloat[MCA-layer 4]{\includegraphics[width=2.7cm]{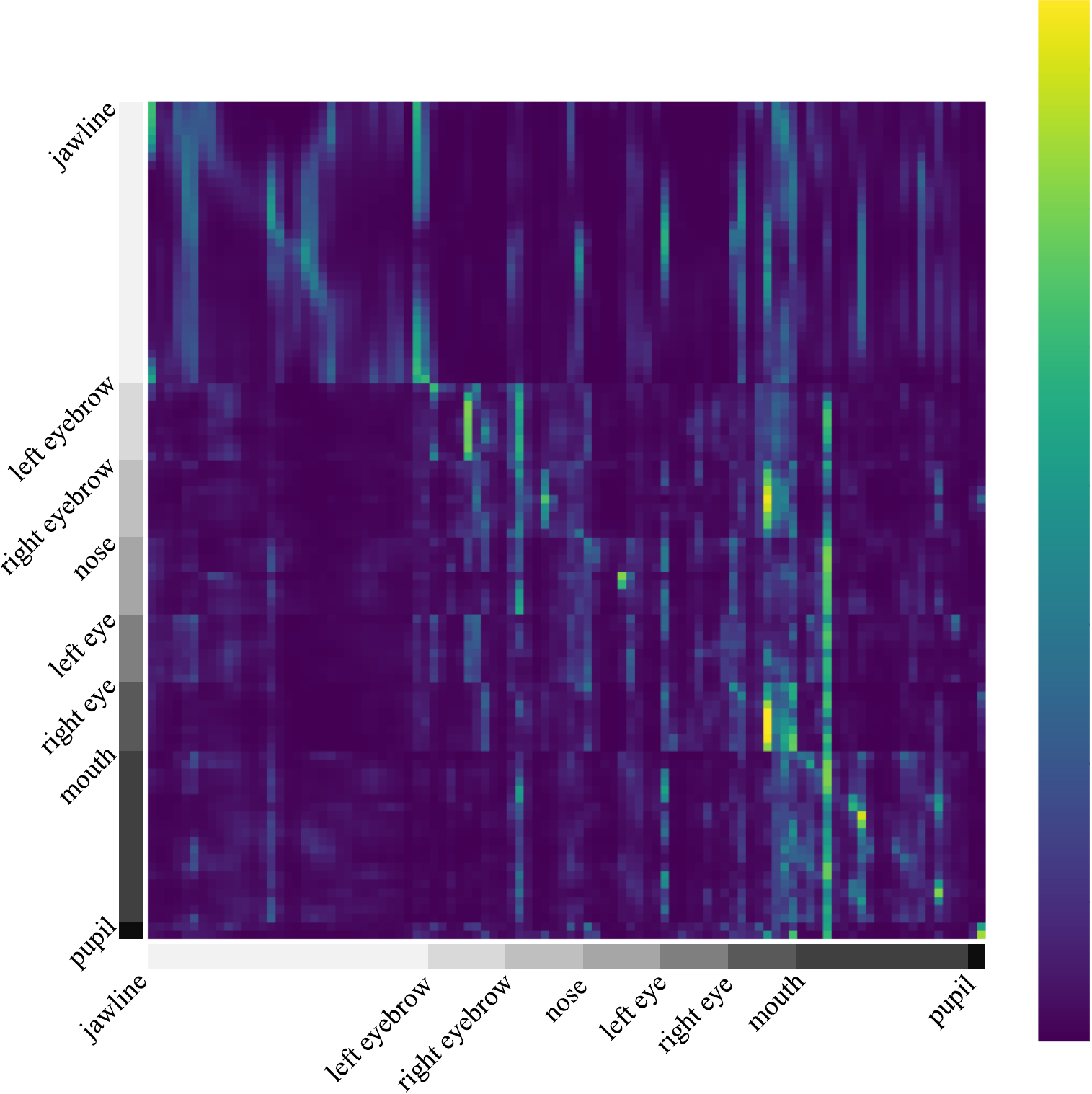}}\hspace{0.1cm}
	\subfloat[MCA-layer 5]{\includegraphics[width=2.7cm]{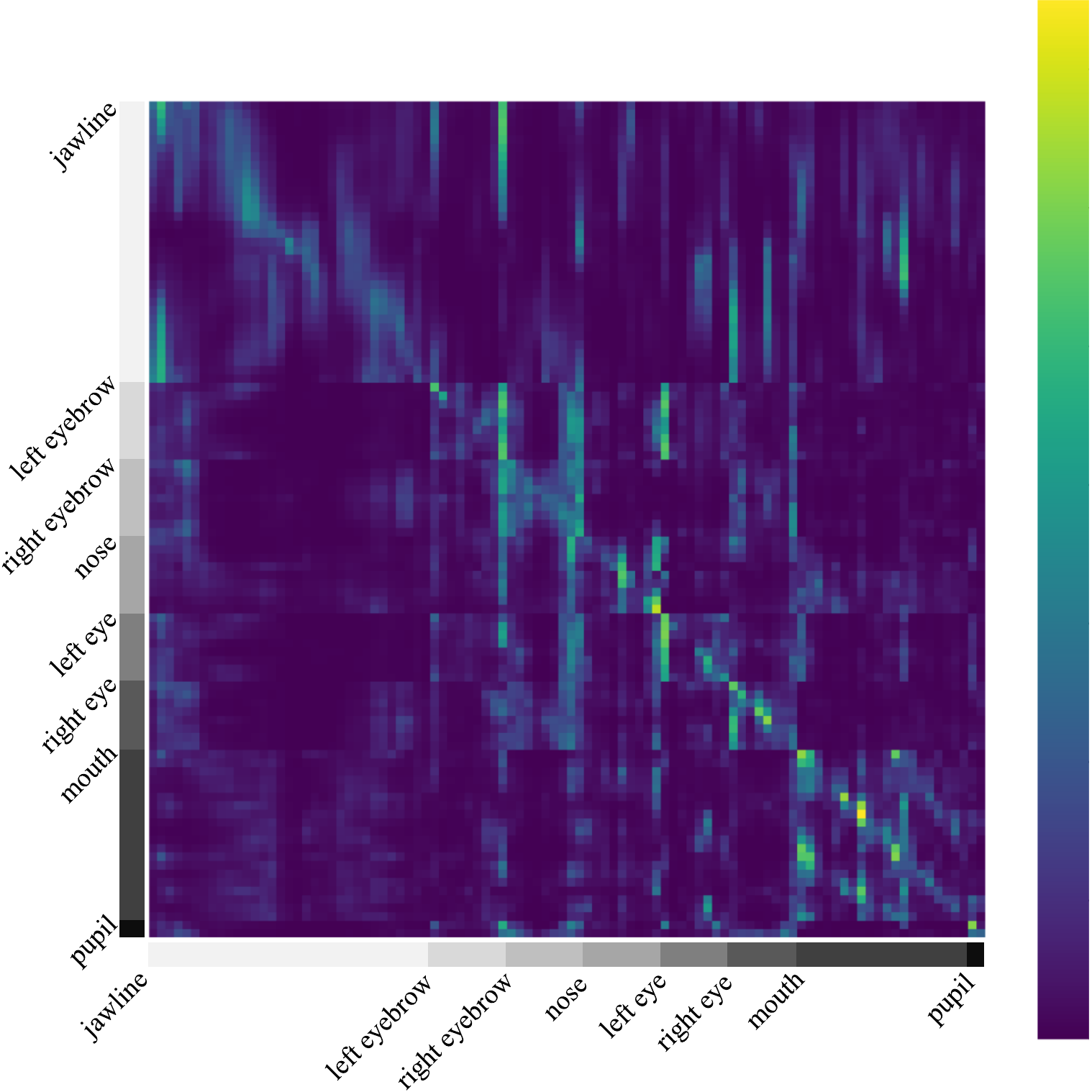}}\hspace{0.1cm}
	\subfloat[MCA-layer 6]{\includegraphics[width=2.7cm]{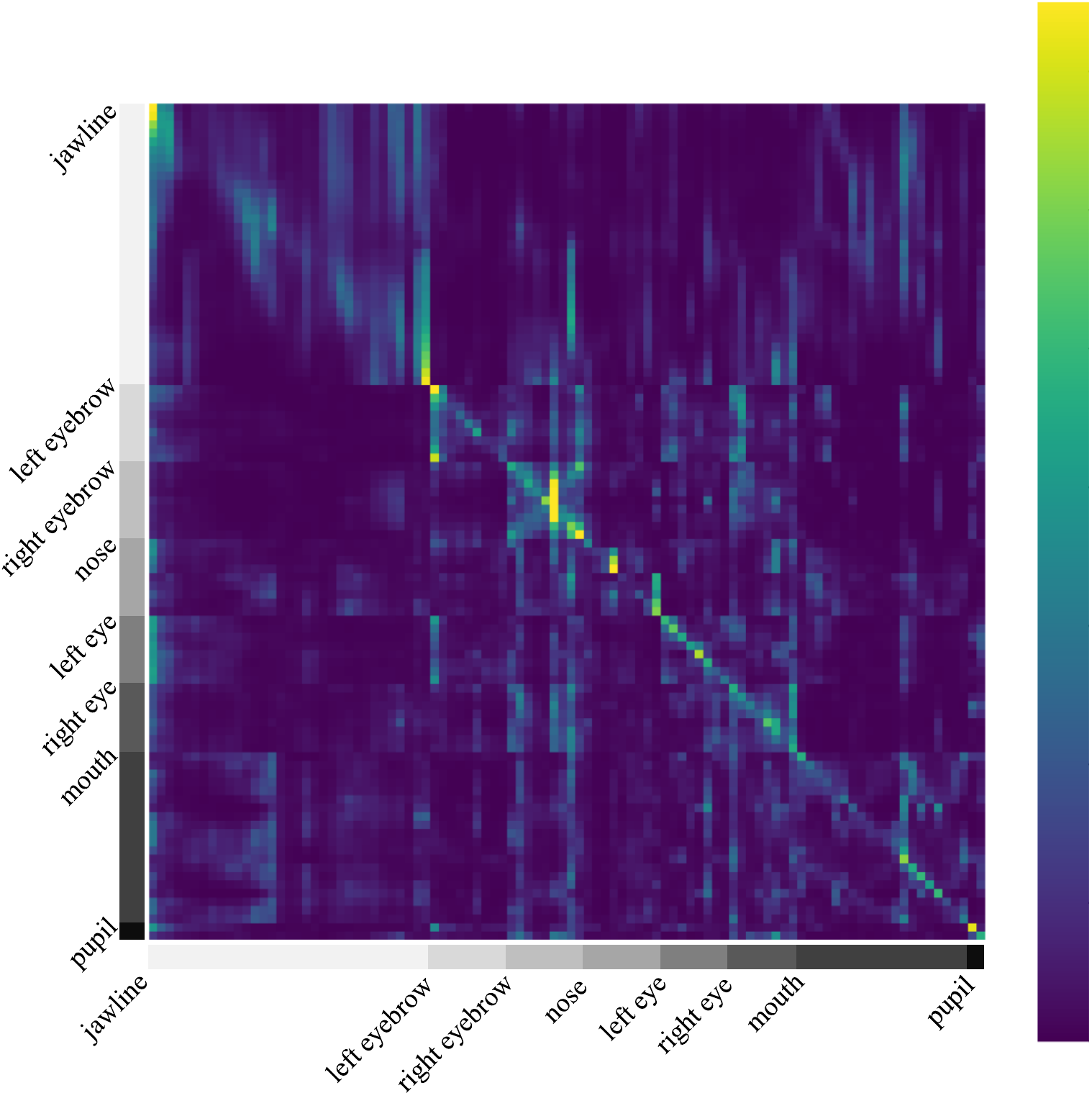}}\hspace{0.1cm}
	\subfloat[MSA-layer 1]{\includegraphics[width=2.7cm]{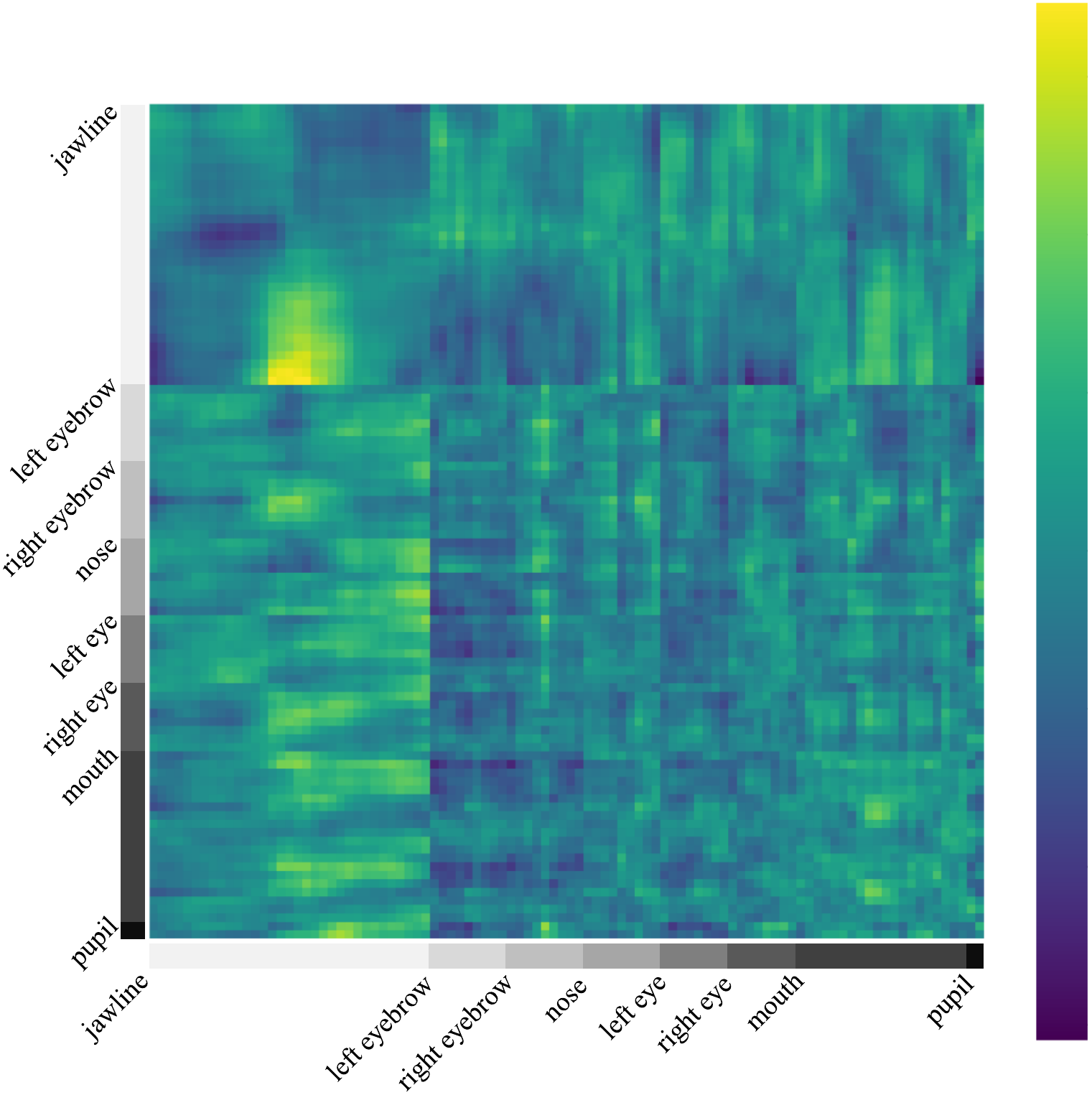}}\hspace{0.1cm}
	\subfloat[MSA-layer 2]{\includegraphics[width=2.7cm]{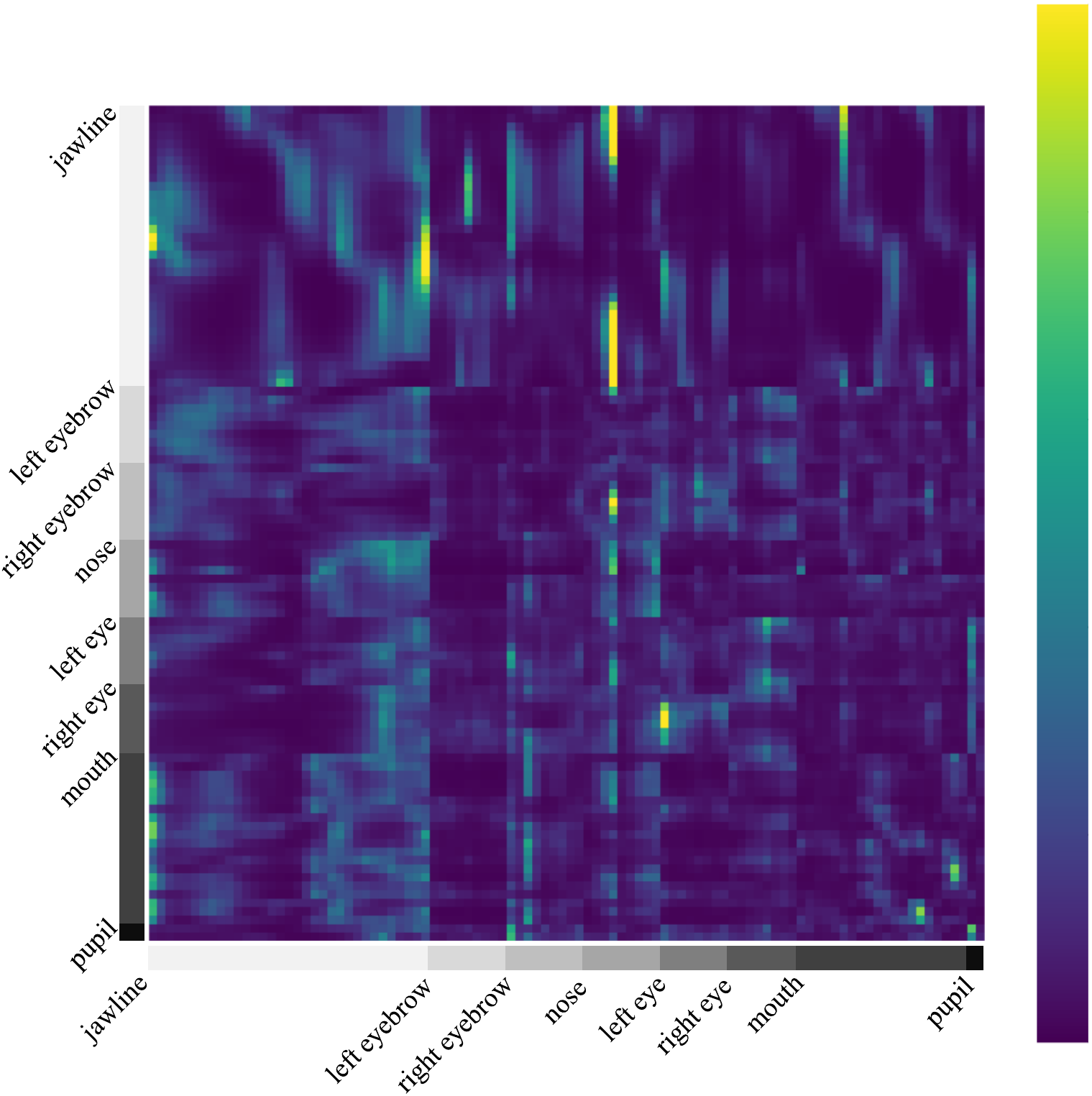}}\hspace{0.1cm}
	\subfloat[MSA-layer 3]{\includegraphics[width=2.7cm]{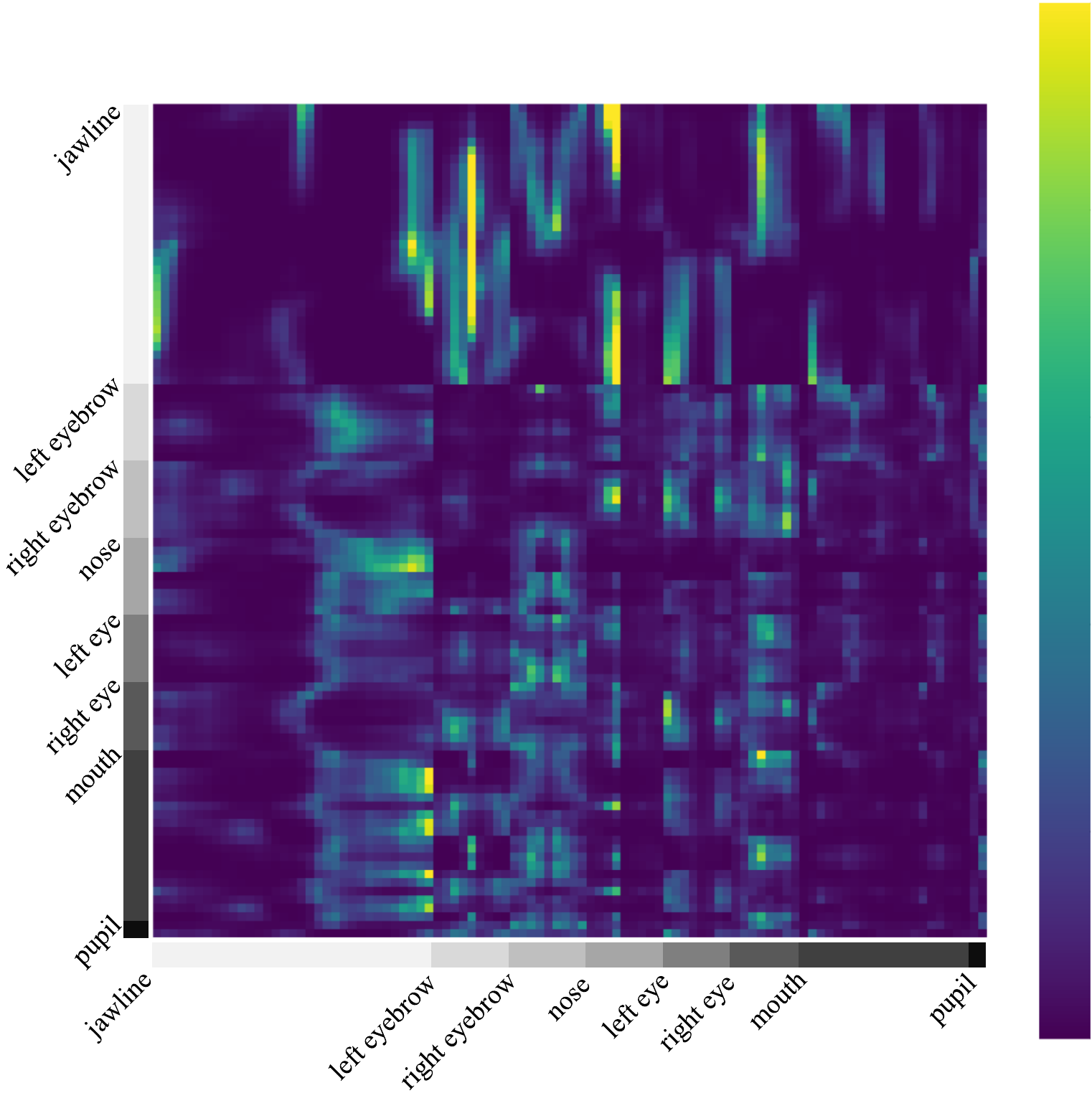}}\hspace{0.1cm}
	\subfloat[MSA-layer 4]{\includegraphics[width=2.7cm]{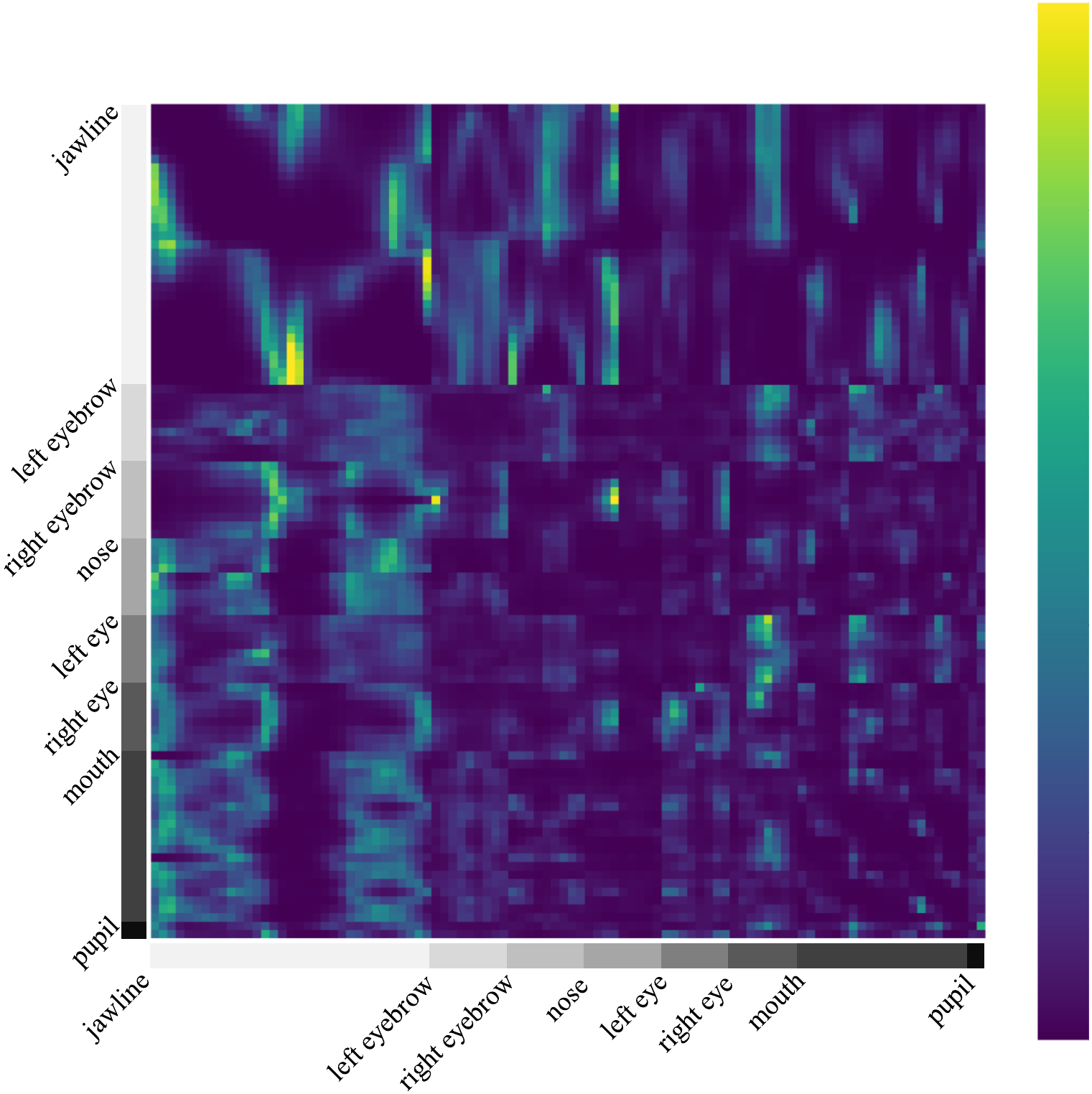}}\hspace{0.1cm}
	\subfloat[MSA-layer 5]{\includegraphics[width=2.7cm]{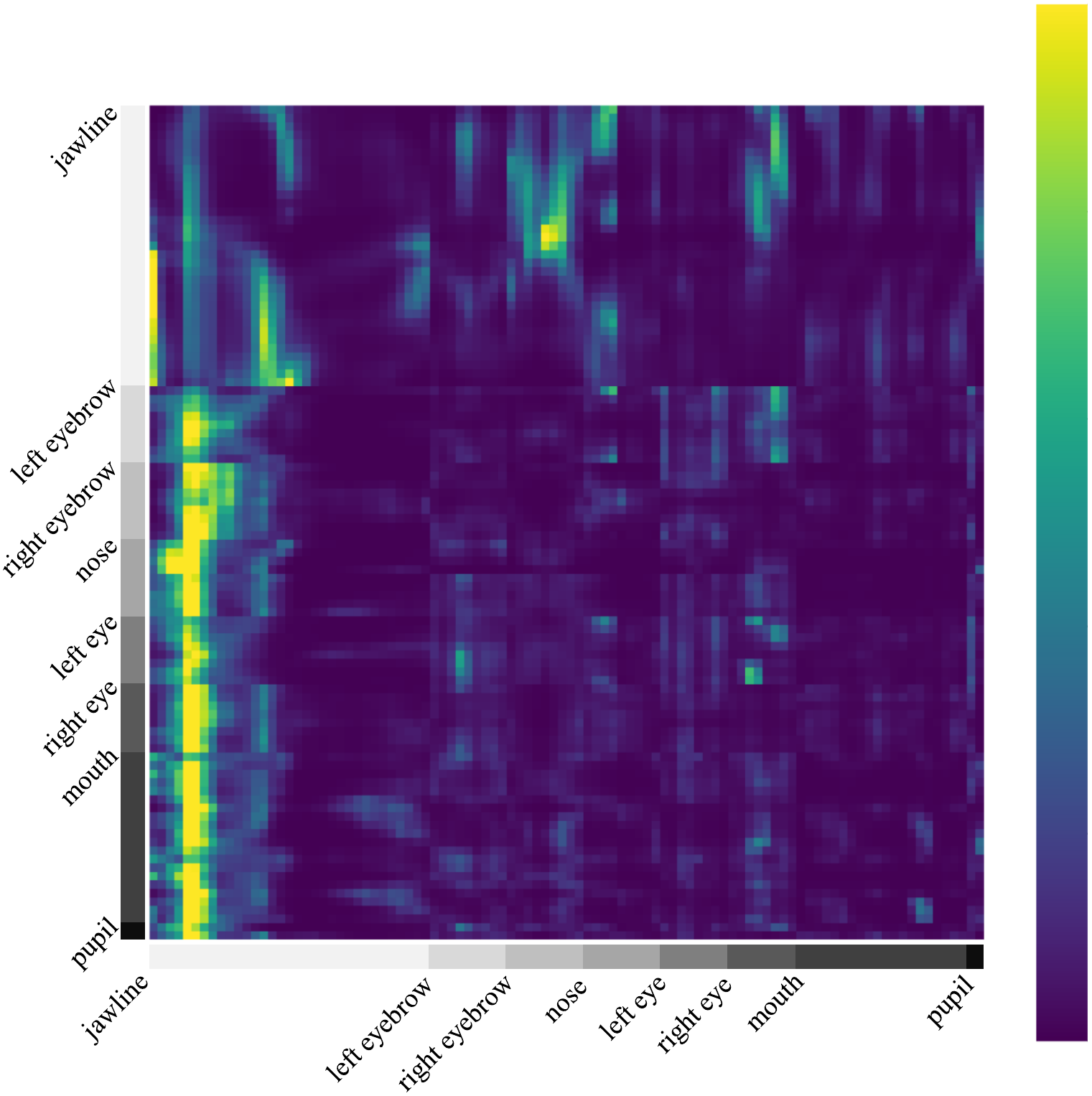}}\hspace{0.1cm}
	\subfloat[MSA-layer 6]{\includegraphics[width=2.7cm]{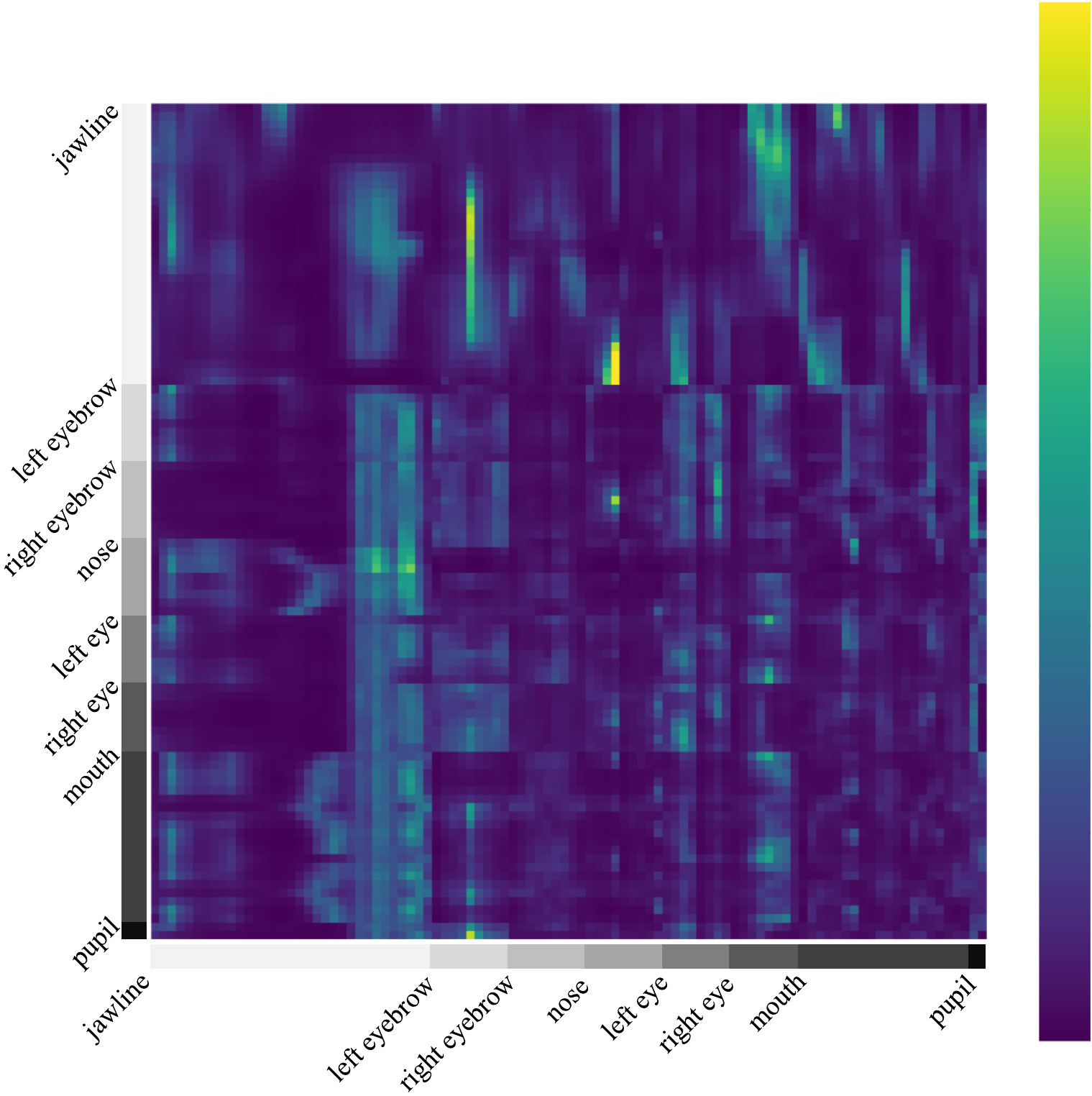}}\hspace{0.1cm}	
	\caption{The statistical attention interactions of MCA and MSA in the final stage on the WFLW test set. Each row indicates the attention weight of the landmark.}
	\label{fig5}
\end{figure*}

\textbf{Evaluation on different coarse-to-fine stages}: to explore the contribution of the coarse-to-fine framework, we train the SLPT with different number of coarse-to-fine stages on the WFLW dataset. The NME, AUC$_{0.1}$ and FR$_{0.1}$ of each intermediate stage and the final stage are shown in Table 5. Compared to the model with only one stage, the local patches in multi-stages model evolve into a pyramidal form, which improves the performance of intermediate stages and final stage significantly. When the stage increases from 1 to 3, the NME of the first stage decreases dramatically from 4.79\% to 4.38\%. When the number of stages is more than 3, the performance converges and additional stages cannot bring any improvement to the model. 

\begin{table}[t!]
	\centering
	\begin{tabular}{m{3.8cm}<{\centering}|m{1.5cm}<{\centering}m{1.5cm}<{\centering}}
		\hline
		Method & FLOPs(G) & Params(M) \\ \hline
		HRNet$^\star$ \cite{HRnet} & 4.75 & 9.66 \\
		LAB \cite{LAB} & 18.85 & 12.29 \\
		AVS + SAN\ \cite{AVS} & 33.87 & 35.02 \\
		AWing \cite{Awing} & 26.8 & 24.15 \\ \hline
		DETR$^\dag$ (98 landmarks) \cite{DETR} & 4.26 & 11.00 \\ 
		DETR$^\dag$ (68 landmarks) \cite{DETR} & 4.06 & 11.00 \\
		DETR$^\dag$ (29 landmarks) \cite{DETR} & 3.80 & 10.99 \\\hline
		SLPT$^\dag$ (98 landmarks) & 6.12 & 13.19 \\
		SLPT$^\dag$ (68 landmarks) & 5.17 & 13.18 \\
		SLPT$^\dag$ (29 landmarks) & 3.99 & 13.16 \\ \hline
	\end{tabular}
	\caption{Computational complexity and parameters of SLPT and SOTA methods. Key: [$^\star$=HRNetW18C, $^\dag$=HRNetW18C-lite]]}
	\label{Tabal8}
\end{table}	

\textbf{Evaluation on MSA and MCA block}: To explore the influence of \textit{query-query} inter relation (eq.1) and \textit{representation-query} inter relation (eq.3) created by MSA and MCA blocks, we implement four different models with/without MSA and MCA, ranging from 1 to 4. For the models without MCA block, we utilize the landmark representations as the queries input. The performance of the four models are tabulated in Table 6. Without MSA and MCA, each landmark is regressed merely based on the feature of the supporting patches in model 1. Nevertheless, it still outperforms other coordinate regression methods because of the coarse-to-fine framework. When self-attention or cross-attention is introduced into the model, the performance is boosted significantly, reaching at 4.20\% and 4.17\% respectively in terms of NME. Moreover, the self-attention and cross-attention can be combined to improve the performance of model further.

\textbf{Evaluation on structure encoding}: we implement two models with/without structure encoding to explore the influence of structural information. With structural information, the performance of SLPT is improved, as shown in Table 7.

\textbf{Evaluation on computational complexity}: the computational complexity and parameters of SLPT and other SOTA methods are shown in Table 8. The computational complexity of SLPT is only $1/8$ to $1/5$ FLOPs of the previous SOTA methods (AVS and AWing), demonstrating that learning inherent relation is more efficient than other methods. Although SLPT runs three times for coarse-to-fine localization, patch embedding and linear interpolation procedures, we do not observe a significant increasing of computational complexity, especially for 29 landmarks, because the sparse local patches lead to less tokens. 

Besides, the influence of patch size and inherent layer number are shown in the Appendix A.4 and A.5.


\subsection{Visualization}
We calculate the mean attention weight of each MCA and MSA block on the WFLW test set, as shown in Fig.5. We find out that the MCA block tends to aggregate the representation of the supporting and neighboring patches to generate the local feature, while MSA block tends to pay attention to the landmarks with a long distance to create the global feature. That is why the MCA block can incorporate with the MSA block for better performance.

\section{Conclusion}
In this paper, we find out that the inherent relation between landmarks is significant to the performance of face alignment while it is ignored by the most state-of-the-art methods. To address the problem, we propose a sparse local patch transformer for learning a \textit{query-query} and a \textit{representation-query} relation. Moreover, a coarse-to-fine framework that enables the local patches to evolve into pyramidal former is proposed to further improve the performance of SLPT. With the adaptive inherent relation learned by SLPT, our method achieves robust face alignment, especially for the faces with blur, heavy occlusion and profile view, and outperforms the state-of-the-art methods significantly with much less computational complexity. Ablation studies verify the effectiveness of the proposed method. In future work, the inherent relation learning will be studied further and extended to other tasks. 

\newpage

{\small
\bibliographystyle{ieee_fullname}
\bibliography{egbib}
}

\newpage
\onecolumn
\section*{Appendix A}

\subsection*{A.1 Contructing Multi-scale Feature Maps for SLPT}

As discussed in Section 4.3, we construct multi-level feature maps for ResNet34, as shown in Fig.6. Supposing the feature map size of $k$-th stage in ResNet34 is $W_{k}\times H_{k}\times d_{k}$, we firstly adopt a $1\times1$ CNN layer to reduce the channels from $d_{k}$ to $C_I/4$. Then, the SLPT crops $N$ patches whose size is $P_{Wk} \times P_{Hk}$ from each level and resizes these patches to $K\times K$. Note that $P_{Wk} \times P_{Hk}$ is $W_{k}/4 \times H_{k}/4$ in the initial coarse-to-fine stage and is reduced by half in each following stage. Finally, the resized patches from different levels are concatenated on the channel dimension which is $C_I$. As the result, the SLPT can utilize both high level and low level features for face alignment.

\begin{figure}[H]
	\centering
	\includegraphics[width=\linewidth]{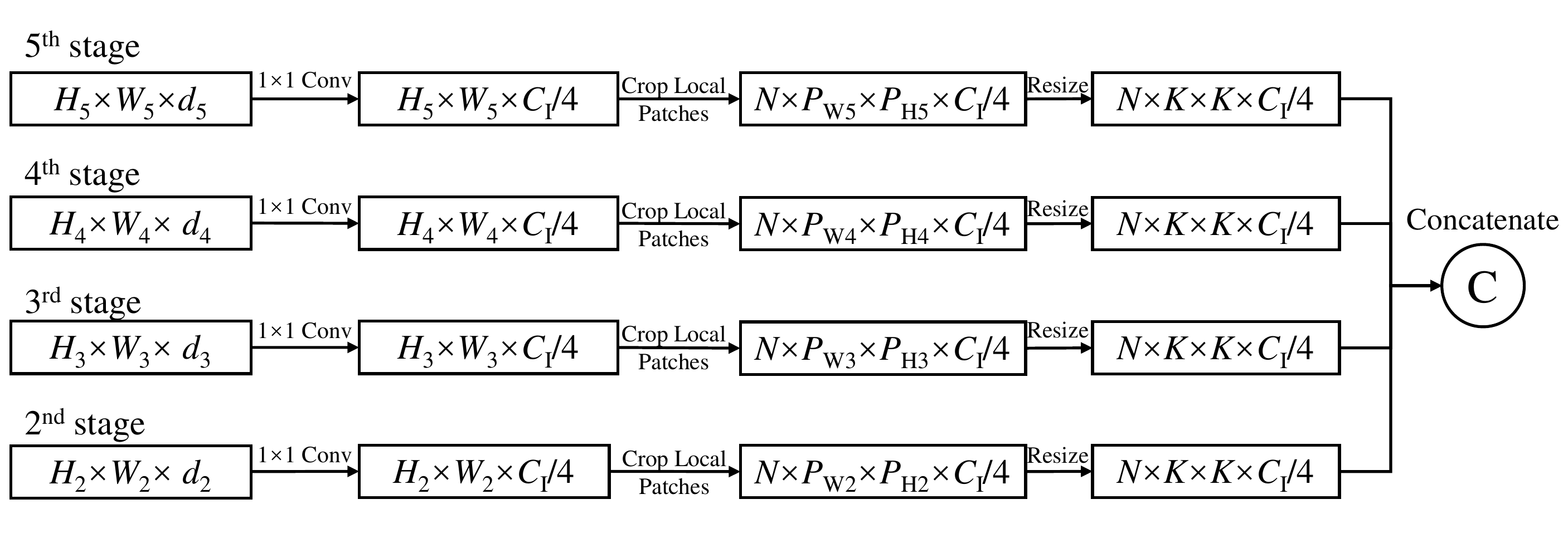}
	\caption{Constructing multi-level feature maps for SLPT}
	\label{fig6}
\end{figure}

\subsection*{A.2 Details of comparison on WFLW}

The comparison results on WFLW test set and its subsets are tabulated in Table 9. SLPT yields the best performance in NME and works at SOTA level on all subsets.



\begin{table}[H]
	\centering
	\begin{tabular}{m{1.7cm}<{\centering}|m{2.6cm}<{\centering}|m{1.2cm}<{\centering}|m{1.2cm}<{\centering}|m{1.4cm}<{\centering}|m{1.6cm}<{\centering}|m{1.3cm}<{\centering}|m{1.3cm}<{\centering}|m{1.2cm}<{\centering}}
		\hline
		Metric    & Method    & Testset & Pose  & Expression & Illumination & Make-up & Occlusion & Blur  \\ \hline
		\multirow{14}{*}{NME(\%)$\downarrow$}
		& LAB \cite{LAB} & 5.27 & 10.24 & 5.51 & 5.23 & 5.15 & 6.79 & 6.32 \\
		& SAN \cite{SAN} & 5.22 & 10.39 & 5.71 & 5.19 & 5.49 & 6.83 & 5.80 \\
		& Coord$^\star$ \cite{HRnet} & 4.76 & 8.48 & 4.98 & 4.65 & 4.84 & 5.83 & 5.49 \\
		& DETR$^\dag$ \cite{DETR} & 4.71 & 7.91 & 4.99 & 4.60 & 4.52 & 5.73 & 5.33 \\
		& Heatmap$^\star$ \cite{HRnet} & 4.60 & 7.94 & 4.85 & 4.55 & 4.29 & 5.44 & 5.42 \\
		& AVS + SAN \cite{AVS} & 4.39 & 8.42 & 4.68 & 4.24 & 4.37 & 5.60 & 4.86 \\
		& LUVLi \cite{LUVLI} & 4.37 & 7.56 & 4.77 & 4.30 & 4.33 & 5.29 & 4.94 \\
		& AWing \cite{AFW} & 4.36 & 7.38 & 4.58 & 4.32 & 4.27 & 5.19 & 4.96 \\
		& SDFL$^\star$ \cite{SCDF}  & 4.35 & 7.42 & 4.63 & 4.29 & 4.22 & 5.19 & 5.08 \\ 
		& SDL$^\star$ \cite{SDL} & 4.21 & 7.36 & 4.49 & 4.12 & 4.05 & {\color{red} $\bm{4.98}$} & 4.82 \\ 
		& HIH \cite{HIH} & {\color{blue} $\bm{4.18}$} & 7.20 & {\color{red} $\bm{4.19}$} & 4.45 & {\color{red} $\bm{3.97}$} & {\color{blue} $\bm{5.00}$} &{\color{blue} $\bm{4.81}$} \\ 
		& ADNet \cite{ADNet} & {\color{red}$\bm{4.14}$} & {\color{red}$\bm{6.96}$} & {\color{blue} $\bm{4.38}$} & 4.09 & 4.05 & 5.06 & {\color{red} $\bm{4.79}$} \\ \cline{2-9}
		& SLPT$^\ddag$ & 4.20 & {\color{blue} $\bm{7.18}$}& 4.52 & {\color{blue} $\bm{4.07}$} & 4.17 & 5.01 & 4.85 \\
		& SLPT$^\dag$ & {\color{red}$\bm{4.14}$} & {\color{red}$\bm{6.96}$} & 4.45 & {\color{red} $\bm{4.05}$} & {\color{blue} $\bm{4.00}$} & 5.06 & {\color{red} $\bm{4.79}$} \\ \hline
		\multirow{14}{*}{FR$_{0.1}$(\%)$\downarrow$}
		& LAB & 7.56 & 28.83 & 6.37 & 6.73 & 7.77 & 13.72 & 10.74 \\
		& SAN & 6.32 & 27.91 & 7.01 & 4.87 & 6.31 & 11.28 & 6.60 \\
		& Coord$^\star$ & 5.04 & 23.31 & 4.14 & 3.87 & 5.83 & 9.78 & 7.37 \\
		& DETR$^\dag$ & 5.00 & 21.16 & 5.73 & 4.44 & 4.85 & 9.78 & 6.08 \\
		& Heatmap$^\star$ & 4.64 & 23.01 & 3.50 & 4.72 & 2.43 & 8.29 & 6.34 \\
		& AVS + SAN & 4.08 & 18.10 & 4.46 & 2.72 & 4.37 &7.74 & 4.40\\
		& LUVLi & 3.12 & 15.95 & 3.18 & {\color{blue} $\bm{2.15}$} & 3.40 & 6.39 & {\color{red} $\bm{3.23}$} \\
		& AWing & 2.84 & 13.50 & 2.23 & 2.58 & 2.91 & 5.98 & 3.75 \\
		& SDFL$^\star$ & {\color{red} $\bm{2.72}$} & 12.88 & {\color{red} $\bm{1.59}$} & 2.58 & 2.43 & {\color{blue} $\bm{5.71}$} & 3.62 \\ 
		& SDL$^\star$ & 3.04 & 15.95 & 2.86 & 2.72 & {\color{red} $\bm{1.45}$} & {\color{red} $\bm{5.29}$} & 4.01 \\
		& HIH & 2.96 & 15.03 & {\color{red} $\bm{1.59}$} & 2.58 & {\color{blue} $\bm{1.46}$} & 6.11 & {\color{blue} $\bm{3.49}$} \\
		& ADNet & {\color{red} $\bm{2.72}$} & {\color{blue} $\bm{12.72}$} & {\color{blue} $\bm{2.15}$} & {\color{blue} $\bm{2.44}$} & 1.94 & 5.79 & 3.54\\ \cline{2-9}
		& SLPT$^\ddag$ & 3.04 & 15.95 & 2.86 & {\color{red} $\bm{1.86}$} & 3.40 & 6.25 & 4.01 \\
		& SLPT$^\dag$ & {\color{blue} $\bm{2.76}$} & {\color{red} $\bm{12.27}$} & 2.23 & {\color{red} $\bm{1.86}$} & 3.40 & 5.98 & 3.88 \\ \hline
		\multirow{14}{*}{AUC$_{0.1}$ $\uparrow$}
		& LAB & 0.532 & 0.235 & 0.495 & 0.543 & 0.539 & 0.449 & 0.463 \\
		& SAN & 0.536 & 0.236 & 0.462 & 0.555 & 0.522 & 0.456 & 0.493 \\
		& Coord$^\star$ & 0.549 & 0.262 & 0.524 & 0.559 & 0.555 & 0.472 & 0.491 \\
		& DETR$^\dag$ & 0.552 & 0.285 & 0.520 & 0.558 & 0.563 & 0.471 & 0.497 \\
		& Heatmap$^\star$ & 0.524 & 0.251 & 0.510 & 0.533 & 0.545 & 0.459 & 0.452 \\
		& AVS + SAN & 0.591 & 0.311 & 0.549 & {\color{red}$\bm{0.609}$} & 0.581 & 0.516 & {\color{red}$\bm{0.551}$} \\
		& LUVLi & 0.557 & 0.310 & 0.549 & 0.584 & 0.588 & 0.505 & 0.525 \\
		& AWing & 0.572 & 0.312 & 0.515 & 0.578 & 0.572 & 0.502 & 0.512 \\
		& SDFL$^\star$ & 0.576 & 0.315 & 0.550 & 0.585 & 0.583 & 0.504 & 0.515 \\ 
		& SDL$^\star$ &  0.589 & 0.315 & 0.566 & 0.595 & {\color{blue}$\bm{0.604}$} & 0.524 & 0.533 \\ 
		& HIH & {\color{blue} $\bm{0.597}$} & 0.342 & {\color{red}$\bm{0.590}$} & {\color{blue}$\bm{0.606}$} & {\color{blue}$\bm{0.604}$} & {\color{blue}$\bm{0.527}$} & {\color{blue}$\bm{0.549}$} \\ 
		& ADNet & {\color{red}$\bm{0.602}$} & {\color{blue} $\bm{0.344}$} & 0.523 & 0.580 & 0.601 & {\color{red}$\bm{0.530}$} & 0.548 \\ \cline{2-9}
		& SLPT$^\ddag$ & 0.588 & 0.327 & 0.563 & 0.596 & 0.595 & 0.514& 0.528 \\
		& SLPT$^\dag$ & 0.595 & {\color{red} $\bm{0.348}$} & {\color{blue}$\bm{0.574}$} & 0.601 & {\color{red} $\bm{0.605}$} & 0.515 & 0.535 \\ \hline
		
	\end{tabular}
	\caption{Performance comparison of the SLPT and the state-of-the-art methods on WFLW and its subsets. The normalization factor is inter-ocular and the threshold for FR is set to 0.1. Key: [{\color{red} \textbf{Best}}, {\color{blue} \textbf{Second Best}}, $^\star$=HRNetW18C, $^\dag$=HRNetW18C-lite, $^\ddag$=ResNet34]}
	\label{Tabal9}
\end{table}

\newpage

\subsection*{A.3 Convergence curves of SLPT and DETR}
The convergence curves of SLPT and DETR is shown in Fig.7. The DETR achieves 4.71\% NME at 391 epochs on WFLW test set. The SLPT achieves better performance with around 8$\times$ less training epochs. With the increasing of training epochs, the performance of SLPT is improved further, achieving 4.14\% NME at 140 epochs.

\begin{figure}[H]
	\centering
	\includegraphics[width=\linewidth]{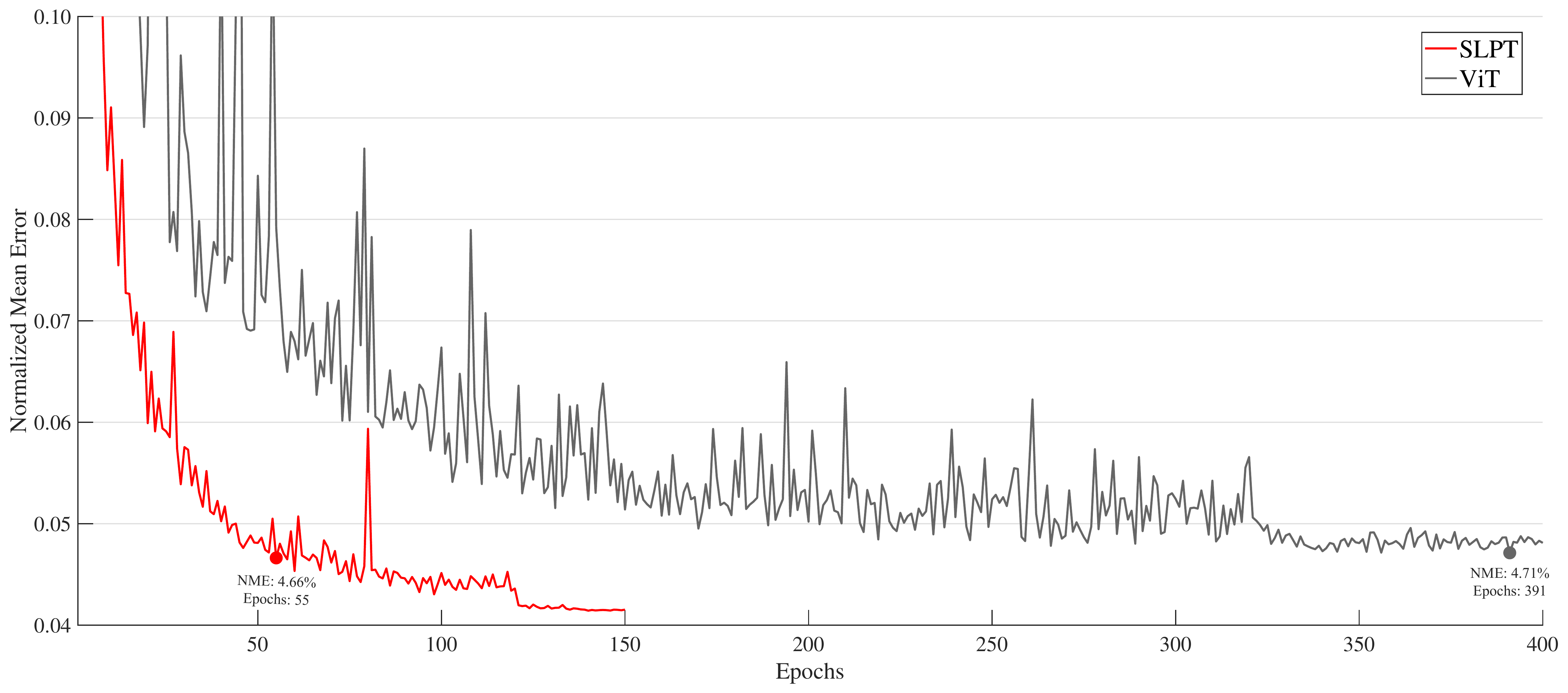}
	\caption{Convergence curves of SLPT and DETR on WFLW test set. The learning rate of SLPT is reduced at 120 and 140 epochs; the learning rate of DETR is reduced at 320 and 360 epochs.}
	\label{fig7}
\end{figure}

\subsection*{A.4 Evaluation on the input patch size}

Each local patch is resized to $K \times K$ and then projected into a vector by a CNN layer with $K \times K$ kernel size. In this section, we explore the influence of the patch size on WFLW test set, as tabulated in Table 10. Compared to $7 \times 7$ patches, the $5 \times 5$ patches lose more information because of the lower resolution, which leads to degradation of the performance. When the patch size is extended from $7 \times 7$ to $9 \times 9$, the parameters of the CNN layer is doubled, which leads to the overfitting on the training set. Therefore, we can also observe a slight degradation with $9 \times 9$ patch size, from 4.14\% to 4.16\% in NME.

\begin{table}[H]
	\centering
	\begin{tabular}{|m{2.0cm}<{\centering}|m{1.2cm}<{\centering}|m{1.2cm}<{\centering}|m{1.2cm}<{\centering}|}
		\hline
		Patch size  & NME(\%) & FR$_{0.1}$(\%) & AUC$_{0.1}$ \\ \hline
		$5 \times 5$ & 4.17\% & {\color{red}$\bm{2.76\%}$} & 0.593 \\ \hline
		$7 \times 7$ & {\color{red}$\bm{4.14\%}$} & {\color{red}$\bm{2.76\%}$} & {\color{red}$\bm{0.595}$} \\ \hline
		$9 \times 9$ & 4.16\% & 2.84\% & 0.594 \\ \hline
	\end{tabular}
	\caption{NME($\downarrow$), FR$_{0.1}$($\downarrow$) and AUC$_{0.1}$($\uparrow$) with different patch sizes $K \times K$ on WFLW test set. Key: [{\color{red} \textbf{Best}}]}
	\label{Tabal10}
\end{table}

\subsection*{A.5 Evaluation on the number of inherent relation layers}

Table 11 demonstrates the influence of inherent relation layer number. The performance of SLPT relies on the inherent relation layer heavily. When the number of inherent relation layers increases from 2 to 12, We can observe a significant improvement, from 4.19\% to 4.12\% in NME. Nevertheless, too many inherent relation layers also increase the parameters and computational complexity dramatically. Considering the real-time capability, we choose the model with 6 inherent relation layers as the optimal model.

\begin{table}[H]
	\centering
	\begin{tabular}{|m{2.0cm}<{\centering}|m{1.2cm}<{\centering}|m{1.2cm}<{\centering}|m{1.2cm}<{\centering}|}
		\hline
		Layer number & NME(\%) & FR$_{0.1}$(\%) & AUC$_{0.1}$ \\ \hline
		2 & 4.19\% & 2.88\% & 0.592 \\ \hline
		4 & 4.17\% & 2.84\% & 0.593 \\ \hline
		6 & 4.14\% & 2.76\% & 0.595 \\ \hline
		12 & {\color{red}$\bm{4.12\%}$} & {\color{red}$\bm{2.72\%}$} & {\color{red}$\bm{0.596}$} \\ \hline
	\end{tabular}
	\caption{NME($\downarrow$), FR$_{0.1}$($\downarrow$) and AUC$_{0.1}$($\uparrow$) with different patch sizes $K \times K$ on WFLW test set. Key: [{\color{red} \textbf{Best}}]}
	\label{Tabal11}
\end{table}

\subsection*{A.6 Further example predicted results and inherent relation maps}
We visualize the predicted results and adaptive inherent relation maps for the samples of COFW, 300W and WFLW, as shown in Fig.8, Fig.9 and Fig.10 respectively. In the inherent relation maps, we connect each point to the point with highest cross-attention weight. The SLPT tends to utilize the visible landmarks to localize the landmarks with heavy occlusion for robust face alignment. For other landmark, it relies more on its neighboring landmark. 

\begin{figure}[H]
	\centering
	\includegraphics[width=\linewidth]{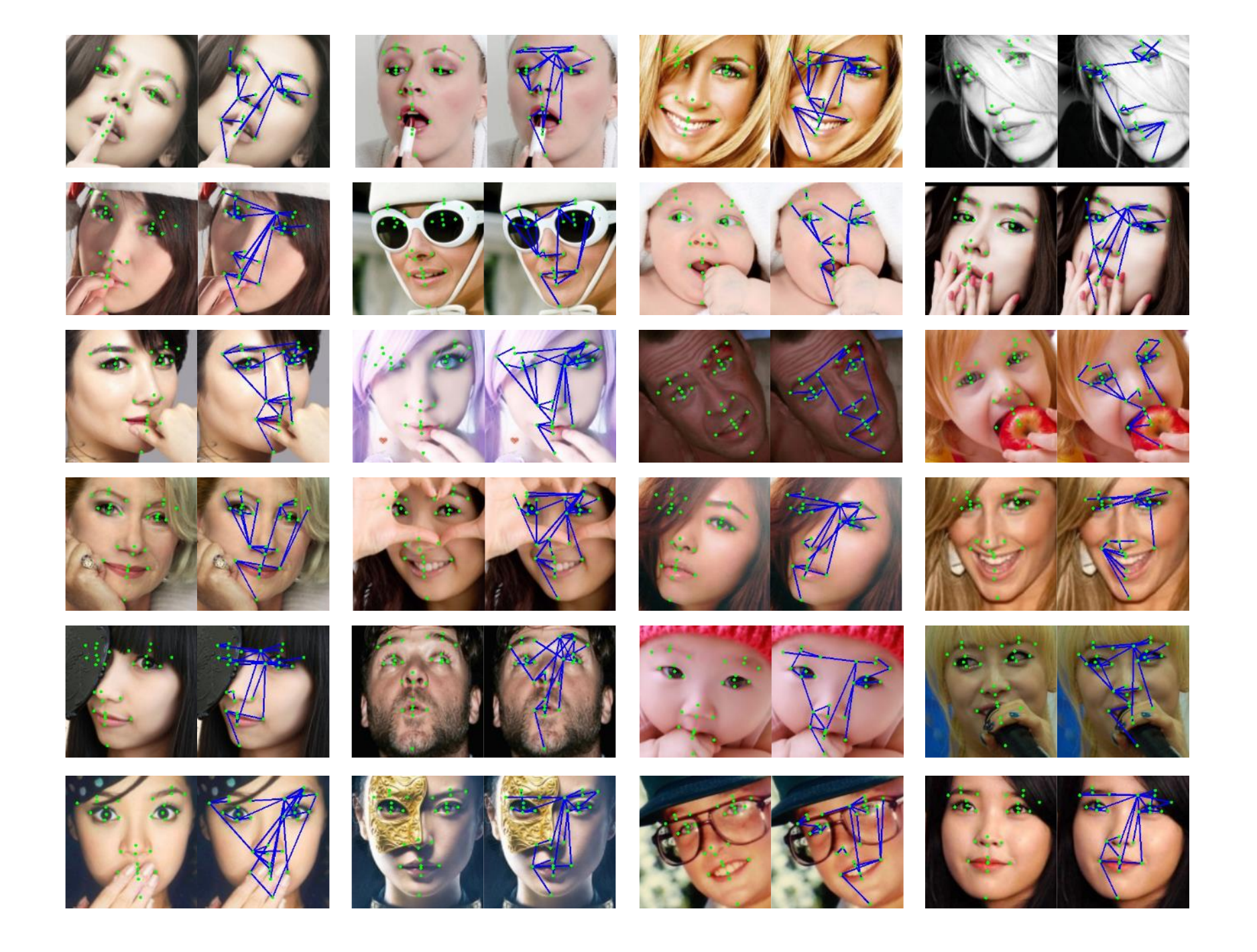}
	\caption{Further example predicted results and attention maps on COFW (random selection)}
	\label{fig8}
\end{figure}

\begin{figure}[H]
	\centering
	\includegraphics[width=\linewidth]{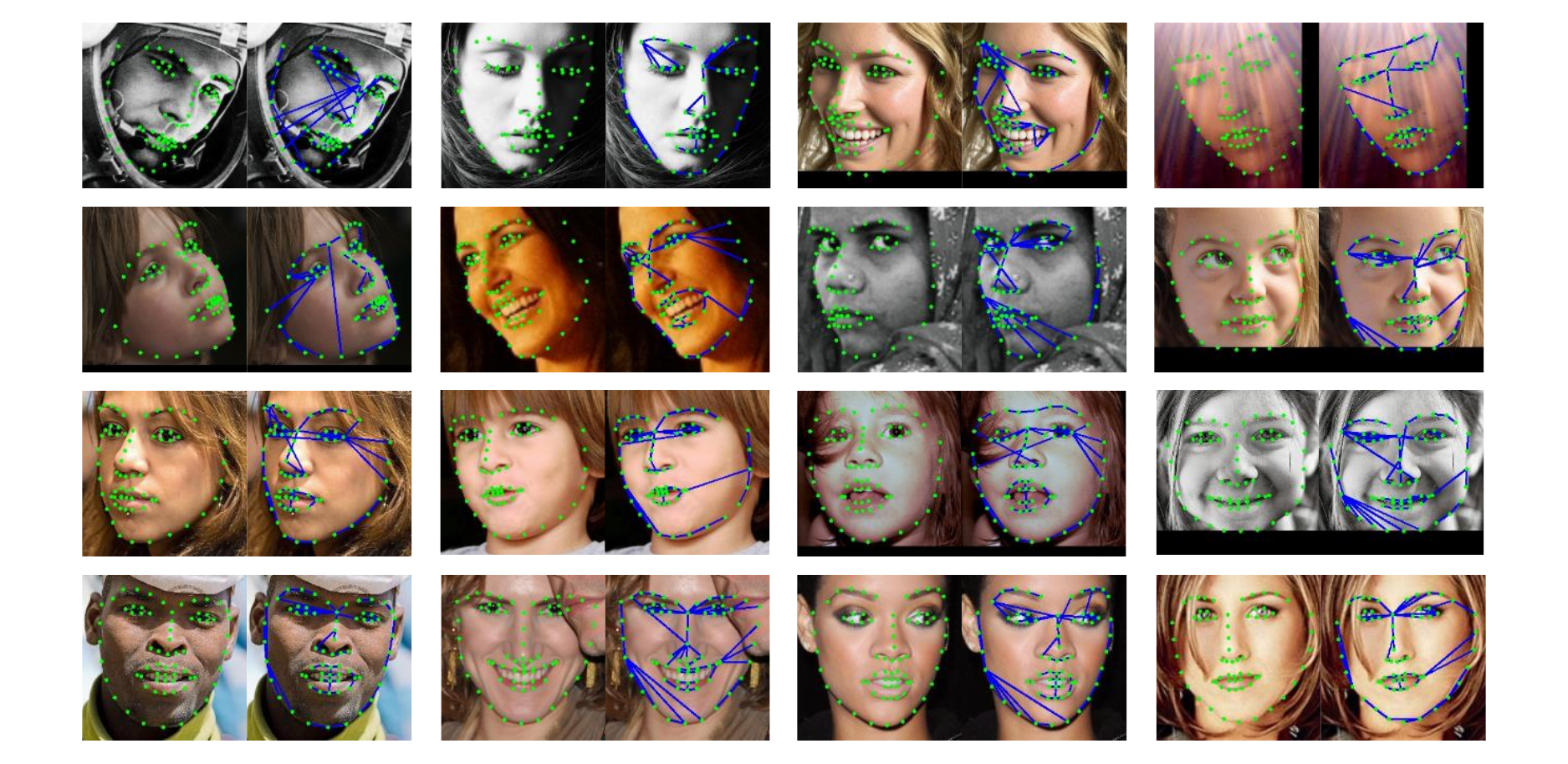}
	\caption{Further example predicted results and attention maps on 300W (random selection)}
	\label{fig9}
\end{figure}

\begin{figure}[H]
	\centering
	\includegraphics[width=\linewidth]{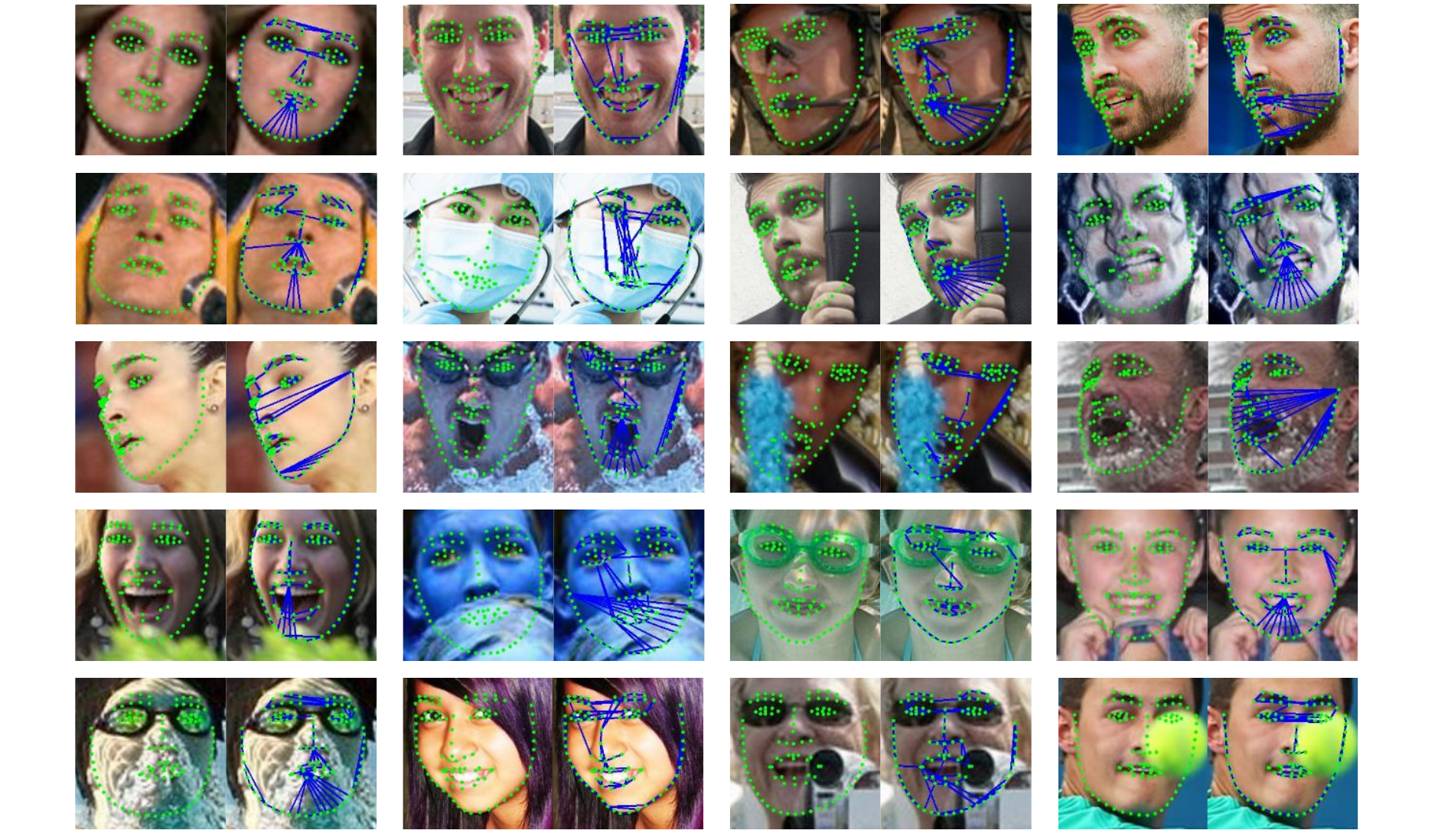}
	\caption{Further example predicted results and attention maps on WFLW (random selection)}
	\label{fig10}
\end{figure}

\end{document}